\documentclass{article}

\usepackage{iclr2027_conference,times}

\usepackage{amsmath,amsfonts,bm}

\def\eqref#1{equation~\ref{#1}}

\def\1{\bm{1}}

\DeclareMathAlphabet{\mathsfit}{\encodingdefault}{\sfdefault}{m}{sl}
\SetMathAlphabet{\mathsfit}{bold}{\encodingdefault}{\sfdefault}{bx}{n}

\usepackage[utf8]{inputenc}
\usepackage[T1]{fontenc}
\usepackage{xcolor}
\usepackage{hyperref}
\hypersetup{
    colorlinks=true, 
    linkcolor=red,
    citecolor=gray, 
    urlcolor=magenta,
    pdftitle={Image Synthesis as an Intermediate for Controllable Time Series Generation},
    pdfsubject={Controllable time series generation},
    pdfkeywords={time series, data augmentation, forecasting, VisualBridge}
}
\usepackage{url}
\usepackage{graphicx}
\usepackage{booktabs}
\usepackage{microtype}
\usepackage{enumitem}
\usepackage{amssymb}
\usepackage{mathtools}
\usepackage{multirow}
\usepackage{wrapfig}

\title{Image Synthesis as an Intermediate for\\Controllable Time Series Generation}

\author{Haochen Yuan\textsuperscript{1}\quad Jing Xie\textsuperscript{1}\quad Yunbo Wang\textsuperscript{1}\thanks{Corresponding author.}\\
\normalfont\small\textsuperscript{1}MoE Key Lab of Artificial Intelligence, Institute of AI,\\
\normalfont\small School of Computer Science, Shanghai Jiao Tong University, Shanghai, China}
\hypersetup{pdfauthor={Haochen Yuan, Jing Xie, Yunbo Wang}}

\usepackage{arxiv-preprint}

\begin{document}

\maketitle

\begin{abstract}
Semantic-driven time-series generation offers a promising way to improve downstream learning in few-shot forecasting, but directly generating numerical sequences from language often fails to preserve the intended temporal structure. We propose \textbf{VisualBridge}, which uses time-series plots as a visual intermediate to bridge high-level temporal semantics and numerical sequences. An MLLM first converts plotted series into structured semantic representations, enabling explicit control over temporal properties such as trend, seasonality, and volatility. We then learn a semantic editing policy with downstream forecasting rewards, allowing the generation process to favor temporal patterns that are beneficial for the target task. The resulting sequences are further modeled by a temporal VAE to produce consistent multivariate augmentations. Experiments on standard public forecasting benchmarks demonstrate that VisualBridge improves few-shot forecasting over conventional augmentation methods, with ablations validating the roles of visual semantic grounding, learned semantic control, and VAE-based generation.

\end{abstract}

\vspace{-4pt}
\section{Introduction}
\label{sec:introduction}
\vspace{-2pt}

Time series generation is a practical tool for data-centric learning, particularly in few-shot scenarios where data scarcity and regime shifts often hinder model generalization. Useful augmentation requires both semantic control and task relevance: generated sequences should reflect specified temporal attributes, such as trend, periodicity, and volatility, while improving downstream learning. Existing approaches range from traditional generative models such as TimeGAN~\citep{yoon2019time} to recent language-model-based forecasting and text-conditioned generation methods~\citep{xue2023promptcast,gu2025verbalts}. Generating a sequence that matches a description, however, does not determine which descriptions yield useful augmentation. This motivates a method that connects semantic generation with downstream feedback.

\begin{figure}[t]
    \centering
    \includegraphics[width=\linewidth]{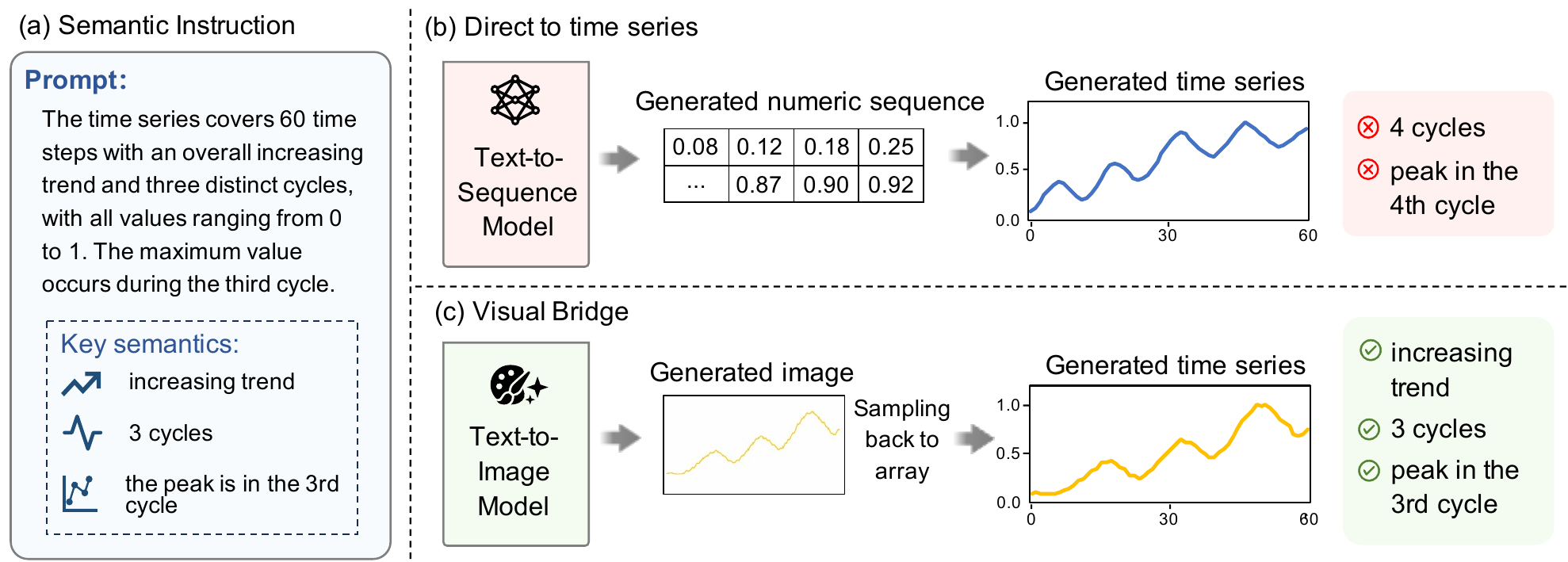}
    \vspace{-16pt}
    \caption{\textbf{Direct numerical generation vs. visual generation for time series.}
    The two right-hand cases use \texttt{GPT-5.6-Luna} for direct numerical generation and \texttt{GPT-image-2} for plot generation, respectively, showing that the visual bridge better preserves the intended temporal semantics.}
    \label{fig:introduction}
    \vspace{-8pt}
\end{figure}

Multimodal large language models (MLLMs) express temporal semantics more reliably through plots than numerical arrays. For instance, a description such as ``an increasing sequence with three fluctuation cycles and the maximum value appearing in the third cycle'' can be rendered with clear periodic structure and peak placement, while direct numerical generation may introduce extra oscillations.
Figure~\ref{fig:introduction} illustrates this case: the visual output better preserves the same semantic instruction than direct numerical generation. Time-series plots make global structure and local variation explicit, providing an intermediate representation between abstract semantics and numerical sequences. This leads to our primary research question: \textit{Can we leverage image generation as a visual intermediate to bridge the gap between abstract semantics and concrete numerical sequences?}

We propose \textbf{VisualBridge}, a semantic time series generation framework that uses image generation as a visual intermediate and controls how generated temporal variations enter downstream training. Given a source window, VisualBridge retains its numerical values for scale recovery and queries an MLLM using its plot and a question-answering (QA) template. The resulting semantic records guide an image generator, and the generated curves are converted back into numerical sequences. The image generator realizes an attribute change together with the record's other temporal structure, without prescribing a separate numerical transformation for each pattern.

To choose useful variations, we incorporate \textbf{downstream-guided semantic control} into VisualBridge. Reinforcement learning (RL) proposes bounded edits to trend, seasonality, and volatility before image generation. Each action is compared with matched no-edit generation using chronological forecasting reward from the training split. Group Relative Policy Optimization (GRPO) trains a semantic action sampler to guide subsequent edits. After training, historical rewards and the frozen policy's context-dependent preferences select edits for fresh generation. Accepted sequences anchor a temporal variational autoencoder (VAE), jointly modeling channels and input--target pairs through reconstruction and forecasting-consistency losses. Forecasting feedback guides semantic changes through the visual pathway before the VAE models the numerical relationships needed for prediction.

Experiments on standard public benchmarks show that VisualBridge improves few-shot forecasting across eight datasets. A same-model comparison holds the generator and semantic records fixed, demonstrating that the visual pathway improves temporal semantic fidelity independently of differences in model capability. Semantic-action evaluation confirms that requested edits translate into directional numerical changes, while component ablations show the benefits of learned semantic selection and VAE modeling. These results support a pipeline that connects semantic generation to forecasting utility through task-guided control and joint numerical modeling.

\vspace{-4pt}
\section{Related Work}
\label{sec:related_work}
\vspace{-2pt}

\paragraph{Semantic-driven time series generation.}

Time series augmentation has been widely used to mitigate data scarcity.
Early methods perturb observed sequences with hand-crafted operations, while learning-based methods model numerical time-series distributions with flows, diffusion, and other generative approaches~\citep{alaa2021generative,yuan2024diffusionts,nikitin2024tsgm}.
Conditional generation provides more explicit control: Time Weaver uses heterogeneous metadata~\citep{narasimhan2024time}, while VerbalTS generates sequences from textual descriptions~\citep{gu2025verbalts}.
VisualBridge also uses semantic conditions, but realizes them through generated plots and curve extraction.
Learning augmentation policies from downstream performance has also been studied in vision, for example in AutoAugment~\citep{cubuk2019autoaugment}.
Our focus is the connection between explicit temporal edits, their visual realization, and forecasting utility, assessed against matched no-edit generations.

\vspace{-8pt}\paragraph{Time series QA.}

Recent work explores temporal reasoning through large language model (LLM)-based forecasting~\citep{jin2023time,gruver2023large}, multi-task time series QA~\citep{kong2025time}, multimodal numerical--textual modeling~\citep{wang2025chattime}, and benchmark-driven evaluation of temporal description and reasoning~\citep{zhou2025cats}.
These studies connect temporal patterns with language, while evaluations also identify limitations in numerical interpretation and time series reasoning~\citep{merrill2024language,arai2025evaluating}.
VisualBridge uses QA to extract structured temporal attributes for generation rather than as the final task.
Numerical recovery and downstream evaluation then assess how these descriptions translate into augmentation samples.

\vspace{-8pt}\paragraph{Time series forecasting.}

Forecasting has evolved from CNNs and RNNs~\citep{torres2021deep,wang2022micn,che2018recurrent,sagheer2019time} to Transformers for long-range temporal dependencies~\citep{li2019enhancing,wu2021autoformer,zhou2021informer,liu2021pyraformer}.
Recent models improve forecasting through decomposition, cross-dimensional modeling, patch-wise tokenization, or variate-wise representation learning~\citep{zhou2022fedformer,zhang2022crossformer,zeng2023transformers,nie2022time,liu2023itransformer}.
Despite these advances, few-shot forecasting remains difficult because a short training prefix often covers only limited temporal regimes.
Existing few-shot forecasting studies address this issue with prior-data-fitted networks, latent-variable formulations, or spatiotemporal adaptation~\citep{dooley2023forecastpfn,jiang2023sequential,xu2024automate,yuan2024lspatiotem}.
Another line uses pretrained time series models or adapts pretrained language models for zero-shot or few-shot forecasting~\citep{das2023decoder,jin2023time,bian2024multi,ttm,lstp2024,yuan2024timer,zhao2024s2ipllm}.
Unlike prior work, our approach focuses on semantic-driven numerical augmentation, using visual semantic grounding and downstream forecasting rewards to generate task-relevant time-series data.

\vspace{-4pt}
\section{VisualBridge for Time Series Generation}
\label{sec:visual_bridge_generation}
\vspace{-4pt}

\subsection{Overview}
\label{sec:mllm_overall_framework}
\begin{figure}[t]
    \centering
    \includegraphics[width=\linewidth]{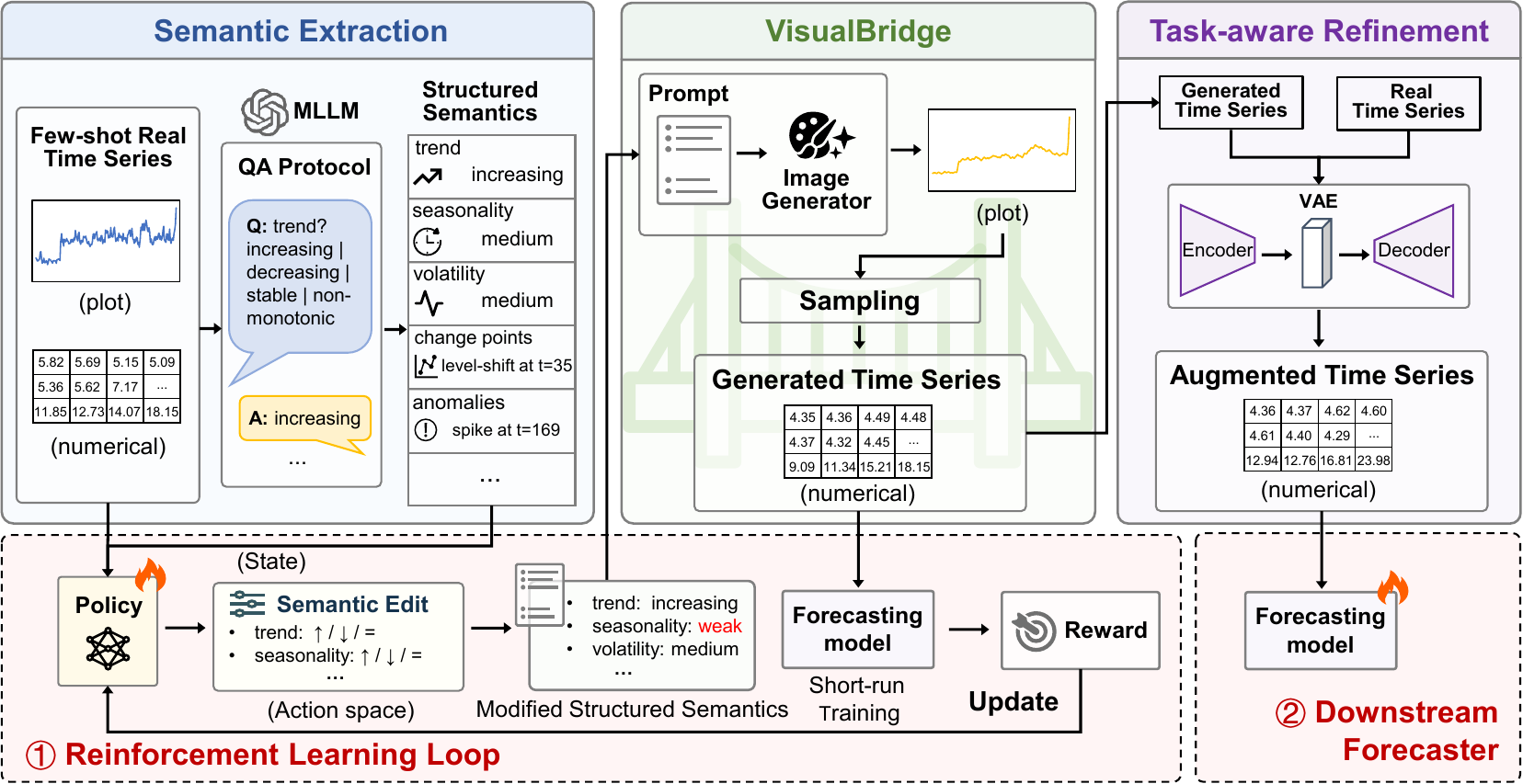}
    \vspace{-18pt}
    \caption{\textbf{Visually-grounded controllable generation pipeline of VisualBridge.}
    Time-series plots connect structured temporal descriptions to numerical generation. An RL policy learns semantic edits using rewards driven by forecasting gains over no-edit generation, measured on held-out probes within the few-shot training split. Edited and no-edit branches use matched initialization and short-run budgets to train forecasters on real and generated data. Generated plots are converted to numerical sequences, and accepted sequences anchor a temporal VAE for multivariate augmentation.}
    \label{fig:overview}
    \vspace{-8pt}
\end{figure}

\vspace{-2pt}
Given a multivariate time series $X \in \mathbb{R}^{T \times C}$, we split it into forecasting windows.
For each channel, a window is written as $s=[x,y]$, where $x$ is the historical input and $y$ is the prediction window.
VisualBridge retains two representations of each window: the numerical sequence, which preserves exact values and scale, and a standardized plot $I_s$, which shows shape, cycles, turning points, and local fluctuations.
Only the plot $I_s$ and a prompt containing window metadata, including identifiers, are given to \texttt{gpt-5.6-luna}, which returns a structured semantic record $a_s$. The numerical sequence $s$ is retained for later calibration and is not included in the semantic-model request.
This record controls \texttt{gpt-image-2} to synthesize a new time-series plot $\tilde{I}_s$ with the same high-level temporal semantics but different local details.
The generated plot is then converted back into a numerical window $\tilde{s}=[\tilde{x},\tilde{y}]$.
The unedited record alone yields a no-edit sequence. Section~\ref{sec:semantic_policy_optimization} adds source-image conditioning for matched action evaluation.
Figure~\ref{fig:overview} shows the full framework; its visual generation path, excluding downstream control and VAE sampling, is
\begin{equation}
    s
    \rightarrow
    I_s
    \rightarrow
    a_s
    \rightarrow
    \tilde{I}_s
    \rightarrow
    \tilde{s}.
    \label{eq:visual_bridge_pipeline}
\end{equation}

For multivariate datasets, VisualBridge applies semantic extraction and visual generation independently to each channel, and then concatenates the recovered channels to form the multivariate window. 
This design focuses the visual bridge on channel-level temporal semantics, such as trend, seasonality, volatility, change points, and anomalies, while the temporal VAE and forecasting backbone model complete multivariate windows.
This channel-wise formulation keeps visual generation and curve extraction stable for high-dimensional datasets, where rendering all variables in one image would impair semantic extraction and numerical recovery.

\vspace{-4pt}
\subsection{Structured Semantic Extraction}
\label{sec:semantic_extraction}
\vspace{-2pt}
For each window, the plot $I_s$ follows a fixed rendering format, including the same window length, axis layout, and line style.
This keeps the visual input comparable across samples and makes visual differences reflect temporal dynamics rather than plotting choices.
The model fills a standardized QA template.
The template records trend direction and strength, seasonality type, period hint and strength, volatility level and change, approximate change points and anomalies, and regime segments.

The output is normalized into a semantic record encoded as JSON $a_s$.
The fixed schema provides consistent attributes for image-generation prompts and subsequent semantic edits.
The full QA schema, output format, and post-processing rules are given in Appendix~\ref{sec:qa_protocol}.

\vspace{-4pt}
\subsection{Semantic-Guided Image Synthesis}
\label{sec:semantic_guided_image_generation}
\vspace{-2pt}
The semantic record $a_s$ is converted into an image-generation prompt and passed to \texttt{gpt-image-2}.
The prompt is built from normalized attributes rather than raw values, so the image generator is guided to preserve coarse temporal semantics such as trend, seasonality, volatility, regime segments, change points, and anomalies, without being asked to trace the original curve.
For each semantic record, sampling from the image generator produces one plot that follows the same semantic constraints while varying in local shape.
The prompt template and plotting configuration are provided in Appendix~\ref{sec:image_to_array}.

\vspace{-4pt}
\subsection{Image-to-Array Conversion}
\label{sec:image_to_series_conversion}
\vspace{-2pt}
The generated plot $\tilde{I}_s$ is recovered with a fixed curve-extraction procedure.
Coordinates within the detected curve bounds are resampled to the required length and mapped to the source window's value range.
We split the sequence into $\tilde{x}$ and $\tilde{y}$ and normalize it as in training.

These sequences realize the extracted records and provide matched controls for evaluating explicit semantic edits.
Section~\ref{sec:semantic_policy_optimization} uses forecasting feedback to learn which edits should guide augmentation.
Appendix~\ref{sec:image_to_array} details curve extraction, coordinate mapping, and interpolation.

\vspace{-4pt}
\section{Downstream-Guided Semantic Control}
\label{sec:semantic_policy_optimization}
\vspace{-2pt}
VisualBridge turns temporal semantics into task-directed augmentation by learning which attribute changes improve forecasting.
A semantic action sampler proposes bounded edits, which the visual pathway converts into numerical candidates for reward evaluation.
The learned preferences and observed rewards select edits for fresh generation; accepted numerical sequences serve as \emph{augmentation anchors} for a temporal VAE.
This separates task-guided semantic selection from joint numerical modeling of the resulting training pairs.

\vspace{-4pt}
\subsection{Semantic Action Sampler}
\label{sec:semantic_intervention}
\vspace{-2pt}
The semantic action sampler is parameterized by a contextual policy $\pi_\theta$.
Let $s_i=[x_i,y_i]$ denote a multivariate source window and $a_i$ its channel-wise semantic records.
Its context combines numerical and semantic descriptors, $h_i=[E_{\mathrm{num}}(s_i),E_{\mathrm{sem}}(a_i)]$.
The policy solves a single-step contextual bandit: it applies one semantic edit to a source batch $\mathcal{B}$ and receives its forecasting reward:
\begin{equation}
    h_{\mathcal{B}}=\frac{1}{|\mathcal{B}|}\sum_{i\in\mathcal{B}}h_i,
    \qquad
    u\sim\widetilde{\pi}_{\theta}(\cdot\mid h_{\mathcal{B}}),
    \qquad
    a_i^{u}=\mathcal{T}(a_i,u),\quad i\in\mathcal{B}.
    \label{eq:semantic_intervention}
\end{equation}
$E_{\mathrm{num}}$ summarizes trend, spectral seasonality, variation scales, and error from a frozen original-only forecaster, also used for candidate checks and VAE training and selection.
$E_{\mathrm{sem}}$ encodes attribute levels and event counts across channels.
These descriptors let action probabilities depend on temporal structure and forecasting difficulty.
During training, $\widetilde{\pi}_{\theta}$ mixes the learned policy $\pi_{\theta}$ with uniform exploration over admissible non-identity edits, sampled without replacement.
Each non-identity action strengthens or weakens trend, seasonality, or volatility by one ordinal level; identity leaves records unchanged.
For example, strengthening medium seasonality changes its strength to strong; image generation realizes this change under the remaining semantic conditions.
Edits are masked unless they change a controlled channel in every batch window; $\mathcal{T}$ repairs dependent fields.
The policy controls ordinal attribute levels, which image generation translates into numerical variations.
The exact descriptors and admissibility masks are specified in Appendix~\ref{sec:semantic_policy_impl}.

\vspace{-4pt}
\subsection{Matched Evaluation of Semantic Actions}
\label{sec:matched_visual_candidates}
\vspace{-2pt}
For each sampled action $u$, the edited record is realized as a numerical candidate:
\begin{equation}
    \tilde{s}_i^{u,r}
    =
    \mathcal{P}\!\left(
    G_{\mathrm{img}}(I_i,a_i^{u};\xi_r)
    \right),
    \label{eq:semantic_visual_candidate}
\end{equation}
Here, $I_i$ contains the source plots, $r$ indexes independent image requests, and $\mathcal{P}$ denotes curve extraction and source-scale calibration.
The identity candidate $\tilde{s}_i^{0,r}$ uses the same reference plots and generation settings as the edited candidate, but retains the unmodified records.
Candidate weights combine curve-extraction quality, source-relative frozen-forecaster error, and edit-direction agreement; identity omits the directional factor.
Identity follows the same visual pathway, providing a reference for the added effect of changing the semantic record.

Candidates undergo numerical, curve-extraction, directional-attribute, and frozen-forecaster checks.
Both edited and identity candidates are blended with the same source window to limit their numerical deviation.
Control is therefore directional within the source scale; its realization before filtering is evaluated in Appendix~\ref{sec:supp_action_realization}.
For high-dimensional series, only a fixed training-selected channel subset is edited; the remaining channels use their aligned no-edit generated values in both branches.
Appendix~\ref{sec:multimodal_control_impl} specifies these checks and the blending rule.

\vspace{-4pt}
\subsection{Training-Only Reward and GRPO}
\label{sec:semantic_reward_grpo}
\vspace{-2pt}
The reward compares an edit with identity on chronological folds drawn entirely from the few-shot training split.
Within each fold, proxies start from the same fresh initialization, independently of the frozen forecaster, and share update budgets and reliability-weighting rules.
A purge gap separates the update prefix from the later probe block.
After updating with action $u$'s candidates, probe mean squared error (MSE) or mean absolute error (MAE), $\ell_{\mathcal{B},f,m}^{u}$, gives the utility relative to identity:
\begin{equation}
    q_{\mathcal{B},f}(u)
    =
    \sum_{m\in\{\mathrm{MSE},\mathrm{MAE}\}}w_m
    \log\frac{\ell_{\mathcal{B},f,m}^{0}+\epsilon}
    {\ell_{\mathcal{B},f,m}^{u}+\epsilon}.
    \label{eq:semantic_downstream_utility}
\end{equation}
Positive utility indicates improvement over no-edit generation; Original measures the net augmentation benefit.
The reward $R_{\mathcal{B}}(u)$ aggregates fold utilities, penalizing instability across folds and late updates, and excess deterioration over identity relative to original-only training.
Capped auxiliary scores measure edit-direction agreement, curve quality, and frozen-forecaster consistency without overriding forecasting utility.
An action with insufficient valid candidates receives a fixed low reward.
The complete reward and acceptance rule are given in Appendix~\ref{sec:reward_grpo_impl}.

GRPO~\citep{shao2024deepseekmath} compares actions evaluated on the same source batch.
For an action group $\mathcal{G}_{\mathcal{B}}$, the advantage is
\begin{equation}
    \widehat{A}_{\mathcal{B}}(u)
    =
    \frac{R_{\mathcal{B}}(u)-\overline{R}_{\mathcal{B}}}
    {\operatorname{Std}_{v\in\mathcal{G}_{\mathcal{B}}}[R_{\mathcal{B}}(v)]+\epsilon}.
    \label{eq:semantic_grpo_advantage}
\end{equation}
Each round updates the policy from an evaluated action group using a clipped likelihood-ratio objective, a reference Kullback--Leibler (KL) divergence penalty, and entropy regularization.
The updated distribution changes subsequent action proposals according to their forecasting reward.
The policy shares parameters across source windows, so each evaluation informs proposals for windows with similar numerical and semantic descriptors.
This lets forecasting feedback guide generation without differentiating through the image generator or curve extraction.
Comparing actions within a source batch makes their relative utility the learning signal, instead of rewarding an action simply because it was evaluated on easier windows.
Identity supplies a fixed control and participates in advantage normalization, but is excluded from the sampled-action objective.
The full training and collection procedure is given in Appendix~\ref{sec:reward_grpo_impl}.

After the allotted rounds, we freeze the policy and collect new candidates from training source windows.
Among admissible edits with positive mean training reward and a qualifying matched evaluation, we maximize mean reward plus weighted log policy probability: the reward summarizes prior evaluations, while the policy preference depends on the current window.
Fresh edited and identity candidates repeat these checks without proxy retraining; accepted edited sequences form $\mathcal{A}$.
Collection excludes search candidates and leaves the policy frozen.

\vspace{-4pt}
\subsection{VAE Modeling and Sampling}
\label{sec:semantic_conditioned_expansion}
\vspace{-2pt}
Channel-wise visual generation specifies temporal shapes, but does not jointly model channels and the input--target relationship.
A temporal convolutional VAE learns these numerical relationships from complete real and generated windows $[x,y]$.
Modeling complete windows preserves the forecasting pair as the unit of generation and lets the decoder coordinate variations across channels.
Value and first-difference reconstruction fit the trajectories and their local increments, while KL regularization shapes the posterior distribution.
With the encoder frozen, we refine the decoder on real training windows using the frozen original-only forecaster's consistency loss on decoded pairs.
This loss encourages the decoded target to be compatible with its decoded history; the reconstruction terms keep the pair close to the training window.
For $b_i\in\mathcal{A}$, we sample from its encoded posterior:
\begin{equation}
    z_{i,k}=\mu_{\phi}(b_i)
    +\gamma\,\sigma_{\phi}(b_i)\odot\varepsilon_k,
    \qquad
    s_{i,k}^{\mathrm{aug}}
    =D_{\eta}(z_{i,k}),
    \label{eq:anchor_vae_sampling}
\end{equation}
Here, $\varepsilon_k\sim\mathcal{N}(0,I)$, $D_{\eta}$ is the decoder, and $\gamma$ controls posterior noise.
Posterior sampling yields alternative realizations around each anchor, preserving its role as the starting point selected through semantic control.
Numerical checks restrict value and variation scales; the frozen forecaster assesses input--target compatibility.
Among valid draws, we retain the pair with the lowest frozen-forecaster MAE; an anchor with no valid draw contributes no sample.
The VAE models joint numerical structure and selects input--target pairs for forecasting.
Appendix~\ref{sec:vae_expansion_impl} gives the implementation. Section~\ref{sec:exp_ablation} evaluates this stage at matched sample counts.

\vspace{-4pt}
\subsection{Forecasting with Controlled Augmentation}
\label{sec:final_forecaster_training}
\vspace{-2pt}
After the policy and VAE are frozen, only retained VAE candidates form $\mathcal{D}_{\mathrm{aug}}$; the augmentation anchors are not added separately.
The final forecaster is trained from scratch with
\begin{equation}
    \mathcal{L}_{\mathrm{final}}
    =
    \mathcal{L}_{\mathrm{ori}}
    +\lambda_{\mathrm{aug}}\mathcal{L}_{\mathrm{aug}},
    \label{eq:final_controlled_augmentation}
\end{equation}
where $\lambda_{\mathrm{aug}}$ controls the contribution of generated supervision.
We select the checkpoint with the lowest validation MSE and evaluate its MAE and MSE on the real test set.
Augmentation is offline and leaves forecasting inference unchanged. Appendix~\ref{sec:final_forecast_impl} gives training and cost details.

\vspace{-4pt}
\section{Experiments}
\label{sec:experiments}
\vspace{-2pt}
\subsection{Experimental Setup}
\label{sec:exp_setup}
\vspace{-2pt}

\paragraph{Datasets and few-shot protocols.}
We evaluate VisualBridge on eight benchmarks: four Electricity Transformer Temperature variants (ETTh1, ETTh2, ETTm1, and ETTm2), along with Weather, Electricity, Traffic, and Exchange, with an input length and prediction horizon of $96$. 
Under our few-shot protocol, only the first $20\%$ of complete forecasting windows in the original training split are used for model fitting and augmentation.
Validation and test splits remain unchanged and are reserved for checkpoint selection and final evaluation, respectively.
All methods and the full-data reference share normalization statistics computed from the standard training interval to maintain consistent error scaling. 
Complete data statistics and split constructions are detailed in Appendix~\ref{sec:dataset_details}.

\vspace{-8pt}
\paragraph{Implementation details.}
We use iTransformer~\citep{liu2023itransformer} as a fixed forecasting backbone.
The semantic QA module uses \texttt{gpt-5.6-luna}, and image generation uses \texttt{gpt-image-2} with a $1536 \times 1024$ output size.
Each source window and channel has one base semantic record.
Proxy iTransformers on chronological, purged training folds provide the paired reward for policy learning.
The final forecaster is trained from scratch on real samples and VAE candidates sampled from the collected augmentation anchors.
Appendices~\ref{sec:qa_protocol} and~\ref{sec:image_to_array} describe QA and generation.
Training details for the backbone, policy learning, VAE sampling, and forecasting are provided in Appendix~\ref{sec:implementation_details}.

\vspace{-8pt}
\paragraph{Compared methods.}
For few-shot forecasting, VisualBridge is compared with the original few-shot setting and representative augmentation baselines:
$\bullet$ Original trains iTransformer only on the available real training samples.
$\bullet$ Gaussian perturbs the original training sequences with additive Gaussian noise.
$\bullet$ Convolve applies convolution-based time series augmentation with \texttt{tsaug.Convolve}~\citep{arundo2020tsaug}.
$\bullet$ TimeGAN trains a generative model with supervised and adversarial objectives~\citep{yoon2019time}.
$\bullet$ Anchor Data Augmentation (ADA) constructs augmented samples through anchor-based transformations~\citep{schneider2024anchor}.

\begin{figure}[t]
    \vspace{-5pt}
    \centering
    \includegraphics[width=\linewidth]{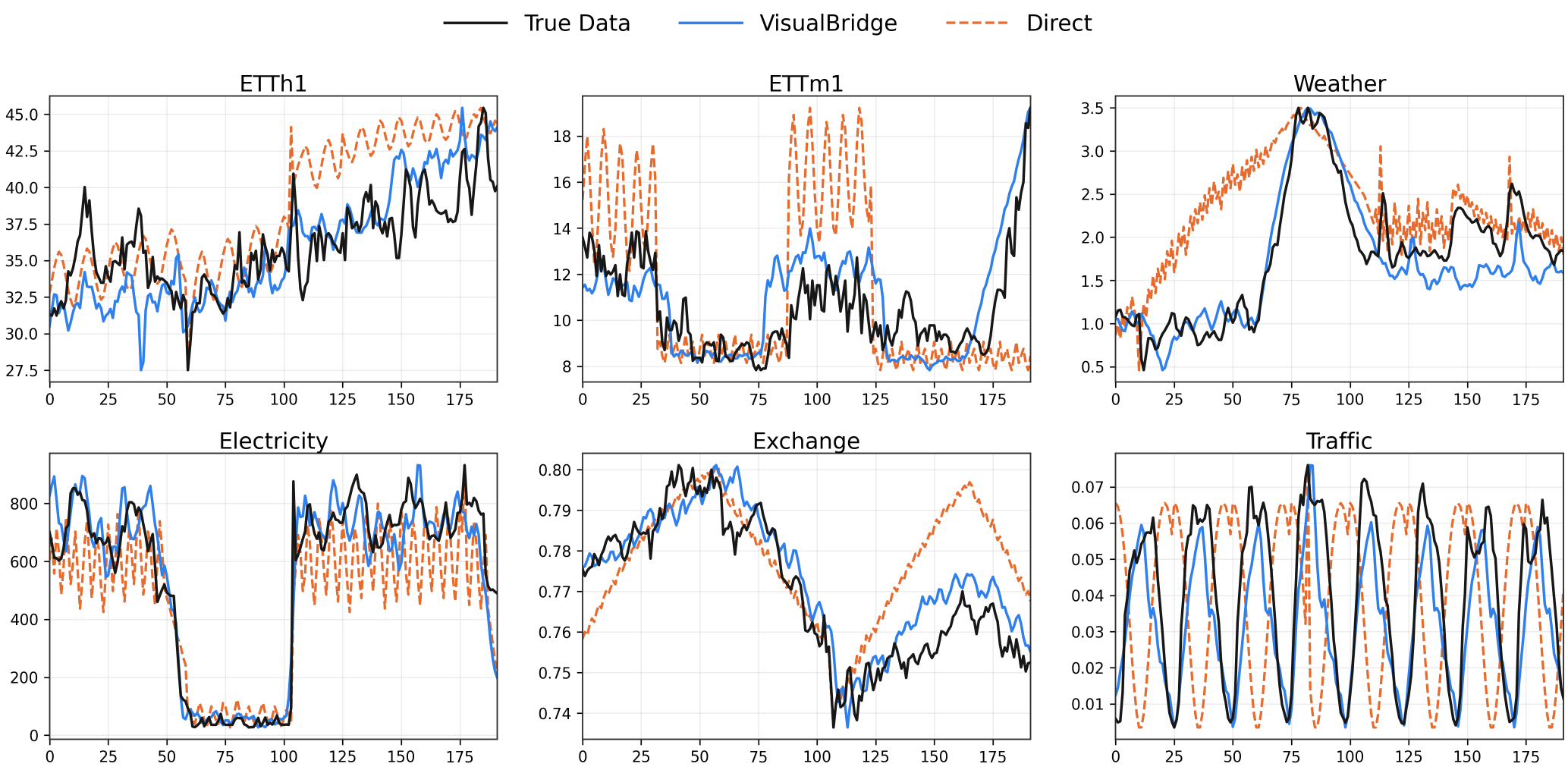}
    \vspace{-18pt}
    \caption{\textbf{Qualitative comparison between direct generation and VisualBridge.}
    Examples use identical, unmodified semantic records across six benchmarks. VisualBridge uses a plot intermediate to preserve recurring patterns and local shape in the recovered numerical sequence.
    }
    \label{fig:VisualBridge_vs_Direct}
    \vspace{-12pt}
\end{figure}

\vspace{-4pt}
\subsection{Fidelity of Time Series Generation}
\label{sec:exp_multimodal_generation}
\vspace{-2pt}

The visual intermediate must carry the temporal information in a semantic description into a numerical sequence before downstream control can select useful variations.
We compare direct language-to-numerical generation (Direct) and VisualBridge conditioned on identical, unmodified semantic records.
Figure~\ref{fig:VisualBridge_vs_Direct} shows how this choice affects generated shape: Direct can capture coarse trends but smooth away fluctuations or displace local events, while the visual pathway better preserves recurring patterns and sharp transitions in these examples.

A semantic description admits multiple numerical realizations, so fidelity concerns the temporal properties it conveys rather than pointwise reproduction of the source.
We measure agreement with the record's seasonality flag using periodicity F1 (Per-F1).
Cycle-count MAE (Cyc-MAE), dominant-frequency error (DF-Err.), and turning-point-rate error (TPR-Err.) complement this semantic measure by testing preservation of the source's repetition count, dominant timescale, and rate of local direction changes.
Peak normalized location error (Peak-NLE) evaluates where the main peak occurs, while permutation-entropy error (PE-Err.) assesses the diversity of local ordering patterns.
Together, these properties capture recurring behavior and local events that a coarse trend alone cannot describe.
The generation settings, evaluation protocol, and metric calculations are given in Appendix~\ref{sec:visual_fidelity_metrics}.

\begin{table}[t]
\centering
\caption{\textbf{Semantic agreement and temporal-structure preservation.}
Direct (D) and VisualBridge (VB) generate from the same semantic records. Per-F1 evaluates the requested seasonality; the remaining metrics compare generated and source structure. Better results within each pair are bold.}
\label{tab:direct_vb_per_dataset}
\setlength{\tabcolsep}{3pt}
\begin{small}
\resizebox{\linewidth}{!}{%
\begin{tabular}{lcccccccccccc}
\toprule
\multirow{2}{*}{Dataset}
& \multicolumn{2}{c}{Per-F1 $\uparrow$}
& \multicolumn{2}{c}{Cyc-MAE $\downarrow$}
& \multicolumn{2}{c}{Peak-NLE $\downarrow$}
& \multicolumn{2}{c}{DF-Err. $\downarrow$}
& \multicolumn{2}{c}{TPR-Err. $\downarrow$}
& \multicolumn{2}{c}{PE-Err. $\downarrow$} \\
\cmidrule(lr){2-3}\cmidrule(lr){4-5}\cmidrule(lr){6-7}\cmidrule(lr){8-9}\cmidrule(lr){10-11}\cmidrule(lr){12-13}
& D & VB & D & VB & D & VB & D & VB & D & VB & D & VB \\
\midrule
ETTh1 & 64.5\% & \textbf{84.3\%} & 3.6190 & \textbf{2.5794} & 0.2468 & \textbf{0.2303} & 0.0381 & \textbf{0.0200} & 0.1731 & \textbf{0.0541} & 0.0981 & \textbf{0.0614} \\
ETTh2 & 62.3\% & \textbf{72.7\%} & 4.0000 & \textbf{2.9032} & 0.2671 & \textbf{0.2239} & 0.0399 & \textbf{0.0240} & 0.2171 & \textbf{0.0768} & 0.1538 & \textbf{0.1331} \\
ETTm1 & 55.5\% & \textbf{62.0\%} & 1.5939 & \textbf{1.4789} & 0.2287 & \textbf{0.2283} & 0.0030 & \textbf{0.0018} & 0.1565 & \textbf{0.0629} & 0.0909 & \textbf{0.0622} \\
ETTm2 & \textbf{69.7\%} & 69.0\% & 1.5699 & \textbf{1.4956} & 0.2233 & \textbf{0.2088} & 0.0060 & \textbf{0.0040} & 0.2003 & \textbf{0.0638} & 0.2263 & \textbf{0.2002} \\
Exchange & 26.3\% & \textbf{27.0\%} & \textbf{1.5161} & 1.5484 & 0.0969 & \textbf{0.0939} & 0.0054 & \textbf{0.0023} & 0.1411 & \textbf{0.0569} & 0.1410 & \textbf{0.0975} \\
Weather & \textbf{39.9\%} & 38.6\% & 1.1993 & \textbf{1.1718} & \textbf{0.1242} & 0.1262 & 0.0052 & \textbf{0.0037} & 0.1577 & \textbf{0.0667} & 0.1823 & \textbf{0.1635} \\
Electricity & 72.2\% & \textbf{93.6\%} & 4.0542 & \textbf{2.7631} & 0.3300 & \textbf{0.2957} & 0.0412 & \textbf{0.0372} & 0.1060 & \textbf{0.0709} & 0.1560 & \textbf{0.1090} \\
Traffic & 82.5\% & \textbf{96.0\%} & 3.9690 & \textbf{3.2860} & 0.2740 & \textbf{0.2247} & 0.0454 & \textbf{0.0366} & 0.1108 & \textbf{0.0705} & \textbf{0.0994} & 0.1034 \\
\bottomrule
\end{tabular}%
\vspace{-15pt}
}
\end{small}
\end{table}

Table~\ref{tab:direct_vb_per_dataset} shows a consistent advantage in dominant-frequency and turning-point-rate errors, indicating better preservation of recurring rhythms and the density of local changes.
The reductions in peak-location and permutation-entropy errors on most datasets extend this evidence to event timing and local pattern complexity.
Per-F1 and Cyc-MAE also generally improve, supporting the realization of recurring behavior conveyed by the semantic record.
The visual intermediate therefore helps preserve the temporal content of a description while allowing variation in its numerical realization, providing a basis for controlled augmentation.
In the same-model comparison in Appendix~\ref{sec:visual_same_model}, VisualBridge also achieves higher Per-F1 and lower Cyc-MAE with the generator held fixed, supporting the visual pathway rather than a difference in model capability.
The forecasting experiments then assess the value of downstream-guided control over these generated variations.

\vspace{-4pt}
\subsection{Few-shot Forecasting Results}
\label{sec:exp_main_results}
\vspace{-2pt}
Table~\ref{tab:main_fewshot} reports few-shot forecasting MAE and MSE across eight benchmarks. VisualBridge achieves the lowest errors on every dataset, outperforming Original and all augmentation baselines. Under a controlled evaluation where all augmentation approaches generate an identical sample volume, conventional baselines (Gaussian, Convolve, TimeGAN, and ADA) exhibit erratic behavior and often degrade predictive accuracy.
This underscores that expanding sample count without structural constraints fails to provide constructive supervision. In contrast, VisualBridge couples visually grounded semantic control with downstream task feedback to generate structurally consistent and task-beneficial temporal patterns.

Beyond absolute accuracy, we report a gap-normalized recovery score $\mathcal{F}$ for few-shot comparisons.
For $\mathcal{M}\in\{\mathrm{MAE},\mathrm{MSE}\}$, the recovery score is defined as
$
    \mathcal{F}_{\mathcal{M}}
    =
    \frac{
    \mathcal{M}_{\mathrm{ori}}
    -
    \mathcal{M}_{\mathrm{aug}}
    }{
    \mathcal{M}_{\mathrm{ori}}
    -
    \mathcal{M}_{\mathrm{full}}
    }
    \times 100\%,
$
where $\mathcal{M}_{\mathrm{ori}}$ is the error of the original model trained without augmentation ($20\%$ training data), $\mathcal{M}_{\mathrm{aug}}$ is the error after augmentation, and $\mathcal{M}_{\mathrm{full}}$ is the error of the same backbone trained with the full training split ($100\%$ training data).
The full-data model is used only as a reference point for computing $\mathcal{F}$.
As shown in Table~\ref{tab:recovery_factor}, VisualBridge is the only framework that achieves strictly positive recovery across every benchmark and metric, substantially outperforming all baselines. Notably, on ETTh2, VisualBridge achieves an MAE recovery factor of $162.4\%$, demonstrating that task-oriented semantic augmentation can effectively regularize downstream representations to surpass the full-data reference under low-resource regimes.

\begin{table}[t]
    \centering
    \vspace{-5pt}
    \caption{\textbf{Main results on few-shot forecasting.}
    Results for stochastic methods are averaged over three generated datasets under a fixed forecasting seed. VisualBridge reports mean $\pm$ standard deviation and consistently outperforms existing data augmentation baselines across all eight public benchmarks.
    }
    \vspace{3pt}
    \label{tab:main_fewshot}
    \setlength{\tabcolsep}{3pt}
    \begin{small}
    \resizebox{\linewidth}{!}{
    \begin{tabular}{lcccccccccccc}
        \toprule
        \multirow{2}{*}{Dataset}
        & \multicolumn{2}{c}{Original}
        & \multicolumn{2}{c}{Gaussian}
        & \multicolumn{2}{c}{Convolve}
        & \multicolumn{2}{c}{TimeGAN}
        & \multicolumn{2}{c}{ADA}
        & \multicolumn{2}{c}{VisualBridge} \\
        \cmidrule(lr){2-3}
        \cmidrule(lr){4-5}
        \cmidrule(lr){6-7}
        \cmidrule(lr){8-9}
        \cmidrule(lr){10-11}
        \cmidrule(lr){12-13}
        & MAE & MSE
        & MAE & MSE
        & MAE & MSE
        & MAE & MSE
        & MAE & MSE
        & MAE & MSE \\
        \midrule
        ETTh1       & 0.459 & 0.472 & 0.463 & 0.478 & 0.465 & 0.477 & 0.471 & 0.480 & 0.460 & 0.472 & \textbf{0.442}\begin{scriptsize}$\pm$0.001\end{scriptsize} & \textbf{0.459}\begin{scriptsize}$\pm$0.001\end{scriptsize} \\
        ETTh2       & 0.353 & 0.307 & 0.356 & 0.308 & 0.358 & 0.311 & 0.357 & 0.314 & 0.359 & 0.318 & \textbf{0.334}\begin{scriptsize}$\pm$0.002\end{scriptsize} & \textbf{0.292}\begin{scriptsize}$\pm$0.001\end{scriptsize} \\
        ETTm1       & 0.416 & 0.428 & 0.414 & 0.427 & 0.404 & 0.433 & 0.407 & 0.421 & 0.406 & 0.426 & \textbf{0.403}\begin{scriptsize}$\pm$0.001\end{scriptsize} & \textbf{0.415}\begin{scriptsize}$\pm$0.002\end{scriptsize} \\
        ETTm2       & 0.268 & 0.190 & 0.269 & 0.190 & 0.272 & 0.193 & 0.273 & 0.192 & 0.270 & 0.192 & \textbf{0.263}\begin{scriptsize}$\pm$0.001\end{scriptsize} & \textbf{0.186}\begin{scriptsize}$\pm$0.001\end{scriptsize} \\
        Weather     & 0.212 & 0.176 & 0.220 & 0.184 & 0.232 & 0.192 & 0.219 & 0.180 & 0.224 & 0.185 & \textbf{0.209}\begin{scriptsize}$\pm$0.001\end{scriptsize} & \textbf{0.173}\begin{scriptsize}$\pm$0.001\end{scriptsize} \\
        Electricity & 0.254 & 0.173 & 0.259 & 0.176 & 0.258 & 0.175 & 0.263 & 0.181 & 0.261 & 0.176 & \textbf{0.249}\begin{scriptsize}$\pm$0.001\end{scriptsize} & \textbf{0.171}\begin{scriptsize}$\pm$0.000\end{scriptsize} \\
        Traffic     & 0.297 & 0.452 & 0.298 & 0.453 & 0.299 & 0.449 & 0.294 & 0.438 & 0.297 & 0.441 & \textbf{0.283}\begin{scriptsize}$\pm$0.001\end{scriptsize} & \textbf{0.425}\begin{scriptsize}$\pm$0.002\end{scriptsize} \\
        Exchange    & 0.213 & 0.091 & 0.214 & 0.093 & 0.211 & 0.087 & 0.211 & 0.088 & 0.220 & 0.100 & \textbf{0.209}\begin{scriptsize}$\pm$0.000\end{scriptsize} & \textbf{0.085}\begin{scriptsize}$\pm$0.000\end{scriptsize} \\
        \bottomrule
    \end{tabular}
    }
    \end{small}
    \vspace{-8pt}
\end{table}

\begin{table}[t]
\vspace{-5pt}
    \centering
    \caption{\textbf{Performance gap recovery measured by $\mathcal{F}$.}
    Percentage of the few-shot-to-full-data error gap recovered by augmentation. Values above $100\%$ surpass the full-data reference; negative values indicate degradation relative to the unaugmented few-shot baseline.
    }
    \vspace{3pt}
    \label{tab:recovery_factor}
    \setlength{\tabcolsep}{3pt}
    \begin{small}
    \begin{tabular}{lcccccccccc}
        \toprule
        \multirow{2}{*}{Dataset}
        & \multicolumn{2}{c}{Gaussian}
        & \multicolumn{2}{c}{Convolve}
        & \multicolumn{2}{c}{TimeGAN}
        & \multicolumn{2}{c}{ADA}
        & \multicolumn{2}{c}{VisualBridge} \\
        \cmidrule(lr){2-3}
        \cmidrule(lr){4-5}
        \cmidrule(lr){6-7}
        \cmidrule(lr){8-9}
        \cmidrule(lr){10-11}
        & $\mathcal{F}_{\mathrm{MAE}}$ & $\mathcal{F}_{\mathrm{MSE}}$
        & $\mathcal{F}_{\mathrm{MAE}}$ & $\mathcal{F}_{\mathrm{MSE}}$
        & $\mathcal{F}_{\mathrm{MAE}}$ & $\mathcal{F}_{\mathrm{MSE}}$
        & $\mathcal{F}_{\mathrm{MAE}}$ & $\mathcal{F}_{\mathrm{MSE}}$
        & $\mathcal{F}_{\mathrm{MAE}}$ & $\mathcal{F}_{\mathrm{MSE}}$ \\
        \midrule
        ETTh1       & -13.0\% & -21.8\% & -19.6\% & -18.1\% & -39.1\% & -29.0\% & -3.3\% & 0.0\% & \textbf{55.4\%} & \textbf{47.2\%} \\
        ETTh2       & -25.6\% & -5.5\% & -42.7\% & -21.9\% & -34.2\% & -38.4\% & -51.3\% & -60.3\% & \textbf{162.4\%} & \textbf{82.3\%} \\
        ETTm1       & 3.4\% & 0.9\% & 20.1\% & -4.3\% & 15.1\% & 6.0\% & 16.8\% & 1.7\% & \textbf{21.8\%} & \textbf{11.1\%} \\
        ETTm2       & -10.5\% & 0.0\% & -42.1\% & -17.9\% & -52.6\% & -11.9\% & -21.0\% & -11.9\% & \textbf{52.6\%} & \textbf{23.9\%} \\
        Weather     & -72.6\% & -94.4\% & -181.6\% & -188.9\% & -63.6\% & -47.2\% & -109.0\% & -106.3\% & \textbf{27.2\%} & \textbf{35.4\%} \\
        Electricity & -26.7\% & -14.6\% & -21.4\% & -9.7\% & -48.1\% & -38.8\% & -37.4\% & -14.6\% & \textbf{26.7\%} & \textbf{9.7\%} \\
        Traffic     & -2.2\% & -1.4\% & -4.4\% & 4.2\% & 6.6\% & 19.5\% & 0.0\% & 15.3\% & \textbf{30.6\%} & \textbf{37.6\%} \\
        Exchange    & -4.9\% & -13.3\% & 9.7\% & 26.6\% & 9.7\% & 20.0\% & -34.1\% & -59.9\% & \textbf{19.5\%} & \textbf{39.9\%} \\
        \bottomrule
    \end{tabular}
    \end{small}
    \vspace{-5pt}
\end{table}

\vspace{-4pt}
\subsection{Ablation Studies}
\label{sec:exp_ablation}
\vspace{-2pt}

We compare six controlled variants:
$\bullet$ \textit{Original} trains only on real few-shot windows.
$\bullet$ \textit{Direct} uses \texttt{gpt-5.6-luna} to generate arrays from unmodified records alone, without images, semantic edits, or VAE sampling.
$\bullet$ \textit{Visual only} uses \texttt{gpt-image-2} and curve extraction from the same records alone, without semantic edits or VAE sampling.
$\bullet$ \textit{Random Semantic} omits policy learning and VAE sampling: proposals are uniform, but historical rewards still select edits during anchor collection.
$\bullet$ \textit{w/o VAE} learns the semantic action sampler with GRPO and uses its collected augmentation anchors directly, omitting VAE sampling.
$\bullet$ \textit{VisualBridge} adds VAE sampling and candidate selection to w/o VAE.
All augmented variants use matched source windows, generated sample budgets, loss weightings, and training batches.

\begin{figure}[t]
    \centering
    \includegraphics[width=\linewidth]{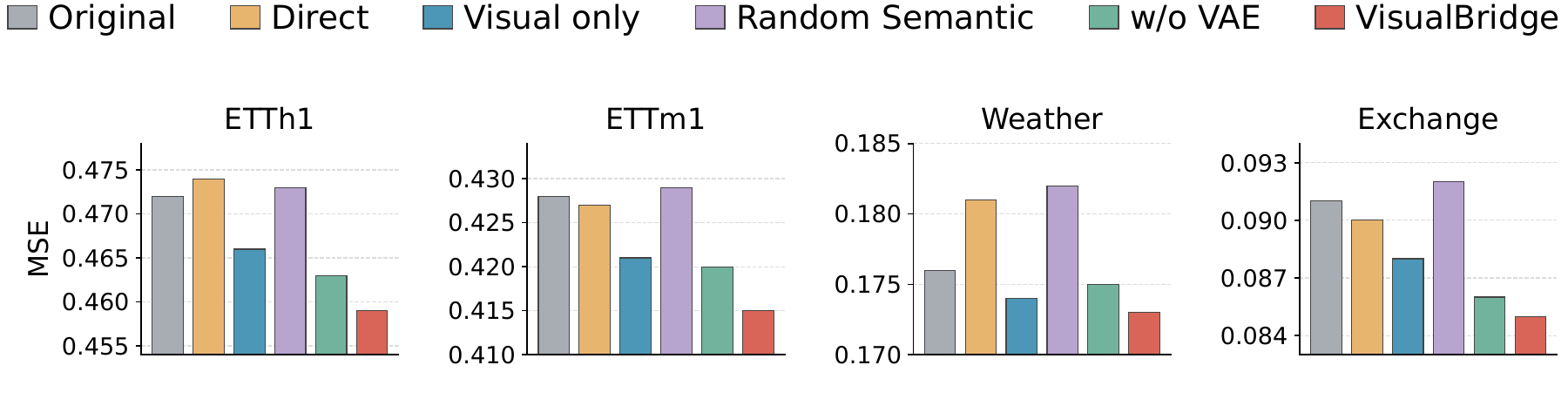}
    \vspace{-25pt}
    \caption{\textbf{Core component ablation by dataset.}
    Bars show mean forecasting MSE. Random Semantic and w/o VAE both omit VAE sampling and differ in whether the semantic action sampler is learned.
    Each panel uses a dataset-specific, nonzero MSE-axis lower bound; error bars are omitted.}
    \label{fig:core_ablation}
    \vspace{-10pt}
\end{figure}

Figure~\ref{fig:core_ablation} compares their forecasting MSE:
First, \textit{Visual only} consistently improves over \textit{Direct} across all datasets, verifying that grounding semantics in visual plots introduces vital geometric and shape-level priors that text-to-numerical generation lacks.
Second, \textit{w/o VAE} steadily outperforms \textit{Random Semantic}, confirming that policy learning guided by downstream rewards actively identifies task-beneficial temporal modifications rather than merely injecting stochastic perturbations.
Finally, \textit{VisualBridge} achieves the lowest MSE on all four benchmarks, outperforming \textit{w/o VAE}. This demonstrates that temporal VAE modeling is essential for regularizing anchor distributions and synthesizing coherent multivariate augmentations.

\vspace{-4pt}
\subsection{Preview of Supplementary Analyses}
\label{sec:exp_supplementary_summary}
\vspace{-2pt}
We supplement the forecasting evaluation with analyses of generation fidelity, semantic control, and transfer.
Appendix~\ref{sec:supp_visual_fidelity} gives metric definitions and complementary diagnostics, including a same-model comparison supporting the visual pathway with the generator held fixed.
Action-realization tests in Appendix~\ref{sec:supp_action_realization} show that semantic edits generally induce the intended numerical changes.
Appendix~\ref{sec:supp_reward_reference} compares matched identity with an original-only reward reference, showing that the former yields augmentation with better forecasting performance.
Experiments with PatchTST and DLinear in Appendix~\ref{sec:supp_backbone_transfer} show that the generated data also improve forecasting beyond the iTransformer used for policy training.

\vspace{-4pt}
\section{Conclusions and Limitations}
\label{sec:conclusion}
\vspace{-2pt}

We presented VisualBridge, a framework that utilizes visual plot representations as an intermediate modality to bridge high-level temporal semantics and numerical sequences. By translating semantic attributes into visual curves, VisualBridge enables controllable temporal generation, while a GRPO-optimized policy actively favors task-beneficial variations guided by downstream forecasting feedback. A temporal VAE further refines candidate sequences into structurally consistent multivariate augmentations. Benchmark experiments confirm that reward-guided semantic synthesis substantially boosts few-shot forecasting performance without modifying downstream architectures.

VisualBridge has two main limitations. First, it focuses on few-shot forecasting, while extending the framework to tasks such as anomaly detection or classification requires adapting the semantic action space and reward criteria. Second, its performance is bound by the generative and perceptual fidelity of underlying multimodal models. Future advancements in visual reasoning and plot synthesis will directly enhance semantic extraction, curve generation, and numerical recovery.

\bibliography{references}
\bibliographystyle{iclr2027_conference}

\appendix
\clearpage
\section*{Appendix}

\newenvironment{vbpromptblock}{%
    \begin{list}{}{%
        \setlength{\leftmargin}{0pt}%
        \setlength{\rightmargin}{0pt}%
        \setlength{\labelwidth}{0pt}%
        \setlength{\labelsep}{0pt}%
        \setlength{\itemindent}{0pt}%
        \setlength{\listparindent}{0pt}%
        \setlength{\topsep}{0pt}%
        \setlength{\partopsep}{0pt}%
        \setlength{\parsep}{0pt}%
        \setlength{\itemsep}{0pt}%
    }%
    \item\relax
    \ttfamily\raggedright
    \setlength{\parindent}{0pt}%
    \setlength{\parskip}{0pt}%
    \setlength{\emergencystretch}{2em}%
}{%
    \par\end{list}
}
\newcommand{\vbpromptline}[1]{\noindent\detokenize{#1}\par}
\newcommand{\vbpromptblank}{\vspace{\baselineskip}}

\section{QA Protocol for Temporal Semantics}
\label{sec:qa_protocol}

VisualBridge uses a structured QA protocol to convert each time series window into an explicit semantic record.
The purpose of the protocol is to reduce the ambiguity of free-form descriptions and provide a stable semantic prompt for time series image generation.
For each window, \texttt{gpt-5.6-luna} receives the corresponding plotted image and a prompt containing metadata (the time-series identifier (ID), dataset, variable, and supplied window length); the numerical sequence is retained for later scale recovery but is not included in the semantic-model request.

\paragraph{QA fields.}
The semantic record consists of a fixed set of fields that describe both global and local temporal properties:

\begin{itemize}[leftmargin=*]
    \item \textbf{Length} is supplied by the preprocessing pipeline rather than estimated from the image. In our experiments it is $192$, and all event and segment positions are approximate integer locations in $[0,191]$.

    \item \textbf{Trend} contains a direction from \texttt{increasing}, \texttt{decreasing}, \texttt{stable}, \texttt{non-monotonic}, and \texttt{unknown}, together with a qualitative strength from \texttt{weak}, \texttt{medium}, \texttt{strong}, and \texttt{unknown}.

    \item \textbf{Seasonality} contains an existence flag, a type from \texttt{fixed}, \texttt{shifting}, \texttt{multiple}, \texttt{irregular}, \texttt{none}, and \texttt{unknown}, a period hint from \texttt{short}, \texttt{medium}, \texttt{long}, and \texttt{unknown}, and a qualitative strength. Here, \texttt{fixed} denotes an approximately stable cycle, \texttt{shifting} denotes a visible change in cycle length or phase, \texttt{multiple} denotes more than one clear cycle, and \texttt{irregular} denotes recurring but non-uniform oscillations. If no seasonality is detected, the type is \texttt{none} and the period hint is \texttt{unknown}.

    \item \textbf{Volatility} contains a level from \texttt{low}, \texttt{medium}, \texttt{high}, and \texttt{unknown}, together with a temporal change from \texttt{stable}, \texttt{increasing}, \texttt{decreasing}, \texttt{regime-shifting}, \texttt{mixed}, and \texttt{unknown}.

    \item \textbf{Change points} are visually salient events with an approximate index and a type selected from \texttt{level\_shift}, \texttt{slope\_change}, and \texttt{variance\_change}.

    \item \textbf{Anomalies} are approximate indexed events of type \texttt{spike}, \texttt{dip}, or \texttt{outlier}, with a qualitative severity from \texttt{low}, \texttt{medium}, and \texttt{high}.

    \item \textbf{Segments} form an ordered, contiguous, non-overlapping partition of the full window. Each segment records its relative level, trend, volatility, and a pattern from \texttt{smooth}, \texttt{oscillatory}, \texttt{weakly oscillatory}, \texttt{irregular}, \texttt{flat}, and \texttt{unknown}.

    \item \textbf{Summary} is a concise natural-language description consistent with the structured fields and limited to visually supported temporal behavior.
\end{itemize}

The schema does not include numerical statistics, normalization parameters, event magnitudes, confidence scores, or numeric control points. Unknown categorical values use \texttt{unknown}, while absent event lists use \texttt{[]} rather than \texttt{null}.

\paragraph{Output format.}
The MLLM response is normalized into a JSON object. The following example shows its structure:
{\scriptsize
\begin{verbatim}
{
  "length": 192,
  "trend": {
    "direction": "non-monotonic",
    "strength": "medium"
  },
  "seasonality": {
    "exists": true,
    "type": "shifting",
    "period_hint": "medium",
    "strength": "strong"
  },
  "volatility": {
    "level": "medium",
    "change": "increasing"
  },
  "change_points": [
    {"index": 70, "type": "level_shift"}
  ],
  "anomalies": [
    {"index": 145, "type": "spike", "severity": "high"}
  ],
  "segments": [
    {
      "start": 0,
      "end": 69,
      "relative_level": "low",
      "trend": "stable",
      "volatility": "low",
      "pattern": "weakly oscillatory"
    },
    {
      "start": 70,
      "end": 191,
      "relative_level": "high",
      "trend": "non-monotonic",
      "volatility": "high",
      "pattern": "irregular"
    }
  ],
  "summary": "A low stable regime shifts upward and becomes volatile."
}
\end{verbatim}
}

\paragraph{Prompt template.}
The prompt used for semantic extraction is:
\begin{vbpromptblock}
\vbpromptline{You are a time-series analyst.}
\vbpromptblank
\vbpromptline{Analyze one plotted time-series image for ts_id={ts_id}, dataset={dataset}, variable={variable}, window_size={window_size}.}
\vbpromptblank
\vbpromptline{Return exactly one valid JSON object and no other text. Do not use markdown fences.}
\vbpromptblank
\vbpromptline{Read only the plotted time-series curve. Ignore the title, axes, grid lines, labels, and other chart decorations.}
\vbpromptblank
\vbpromptline{Set length to exactly {window_size}; it is provided metadata, not a visual estimate.}
\vbpromptblank
\vbpromptline{All index, start, and end values are approximate horizontal locations inferred from the curve and must be integers from 0 to {window_size-1}.}
\vbpromptblank
\vbpromptline{Use only the enum values shown in the schema. Use "unknown" when a categorical property cannot be determined. Never use null.}
\vbpromptblank
\vbpromptline{Use [] when there are no change points or anomalies. Report only visually salient events; do not label ordinary fluctuations as events.}
\vbpromptblank
\vbpromptline{Segments must be ordered, contiguous, non-overlapping, and together cover the full range 0 through {window_size-1}. Create a new segment only for a meaningful level, trend, volatility, or pattern regime.}
\vbpromptblank
\vbpromptline{If no regime boundary is visible, return one full-window segment.}
\vbpromptline{If seasonality.exists is false, set type to "none" and period_hint to "unknown".}
\vbpromptblank
\vbpromptline{Do not add statistics, normalization values, magnitudes, confidence scores, control points, or keys not present in the schema.}
\vbpromptblank
\vbpromptline{Keep the summary concise and consistent with the structured fields.}
\vbpromptline{Use this schema and keep keys unchanged:}
\vbpromptline{{semantic_schema}}
\end{vbpromptblock}

\paragraph{Post-processing.}
The raw MLLM output is parsed and normalized before being used as an image generation prompt.
Invalid or incomplete responses are handled by extracting the JSON object from the raw text, enforcing the expected length, normalizing categorical values, clamping and deduplicating event indices, replacing malformed event lists with empty lists, and reconstructing the segments as contiguous full-window coverage. If no usable segment is returned, a single full-window segment with unknown attributes is inserted. Responses that do not contain any parseable JSON object are retried by the pipeline.
The first successfully parsed and normalized response defines the semantic record for each source window and channel.
This normalization step ensures that the image generation model receives a consistent semantic description across datasets and samples.

\paragraph{Admissible semantic edits.}
The extracted record is retained as an immutable reference, and the policy operates on an action-specific copy.
An action either leaves the record unchanged or applies a bounded strengthen/weaken edit to one root attribute among trend, seasonality, and volatility.
The available actions depend on the current record: edits that cannot change the attribute or would violate the schema are masked before sampling.
After an edit, dependent fields are normalized under the same rules as the QA output; for example, seasonality type and period are cleared when seasonality is removed, while segments and the summary are made consistent with the edited root fields.
The source identifier, action, original record, and edited record are retained together for every generated candidate.

\section{Image Generation and Image-to-Array Conversion}
\label{sec:image_to_array}

We detail the plotting, prompting, and curve-extraction procedure introduced in Section~\ref{sec:visual_bridge_generation}.
Each supervised window has a length of $192$, with an input window of $96$ and a prediction window of $96$.
The generated plot is converted under the same normalization protocol and then split into $[\tilde{x},\tilde{y}]$ for supervised forecasting.

\paragraph{Plotting configuration.}
Source windows use a shared plotting style: a $960 \times 360$ pixel canvas at $120$ dots per inch (dpi), with a blue line (\texttt{\#1f77b4}, width $1.4$), labeled axes, and a light grid.
The axes represent within-window time and the observed value, with the vertical range set by each window.
This local scale makes temporal shape visible across variables with different units.
For multivariate datasets, VisualBridge processes each channel independently: plotting, semantic extraction, generation, and curve extraction produce univariate sequences, which are concatenated to reconstruct the multivariate window.

\paragraph{Image generation prompt.}
The image generation prompt inserts the semantic JSON into the following template:
\begin{vbpromptblock}
\vbpromptline{Generate a data-extraction-friendly time-series line image from the semantic JSON below.}
\vbpromptline{The image will be parsed by a color-based curve extractor, so visual precision matters more than style.}
\vbpromptblank
\vbpromptline{Canvas and geometry requirements:}
\vbpromptline{1) White background only.}
\vbpromptline{2) Exactly one continuous polyline.}
\vbpromptline{3) Main line color must be #1f77b4; no gradients, shadows, glow, or texture.}
\vbpromptline{4) Line width 4 px to 6 px.}
\vbpromptline{5) No title, no legend, no axis ticks, no numeric labels, no text, no watermark.}
\vbpromptline{6) No grid lines and no border box; the blue line should be the only non-white object.}
\vbpromptline{7) Keep at least 8 percent white margin on all sides; never crop the line.}
\vbpromptline{8) Follow the supplied trend, seasonality, volatility, change points, anomalies, and segments.}
\vbpromptline{9) Place any listed event at its specified time-series index.}
\vbpromptline{10) Do not add extra spikes unless they are listed as anomalies.}
\vbpromptline{ts_id={ts_id}}
\vbpromptline{semantic_json={semantic_json}}
\end{vbpromptblock}
In our implementation, the image generation backend uses \texttt{gpt-image-2} with an output size of $1536 \times 1024$ and the \texttt{low} quality setting, and requests one image for each semantic record evaluated by the policy.
For policy evaluation and augmentation-anchor collection, the source plot is supplied as a reference image for both the edited and identity requests.
The geometry instructions above are shared; the edited request substitutes its modified semantic JSON, while identity retains the original JSON.
These matched reference-image requests are distinct from the text-only plot generation used to assess the visual pathway independently of RL.

\paragraph{Curve extraction.}
We isolate the blue trace using a fixed color mask. For color intensities $R,G,B$, a pixel is retained when
\[
    B\ge80,\qquad B-R\ge30,\qquad B-G\ge10,\qquad
    B-\frac{R+G}{2}\ge20.
\]
Within their bounding box, the curve position is the median vertical coordinate in each column.
Missing columns are filled by linear interpolation, with nearest-boundary values used outside the detected range.

We invert the vertical coordinates, which increase downward, and linearly resample the curve to $192$ points.
Let $z_t$ denote the resulting trajectory.
Its values are mapped to the reference window's range $[a,b]$:
\[
    \tilde{s}_t=a+\frac{z_t-z_{\min}}{z_{\max}-z_{\min}}(b-a).
\]
Here, $z_{\min}$ and $z_{\max}$ are the extrema after resampling; a constant curve is assigned the midpoint of $[a,b]$.
This mapping fixes the numerical range while retaining the generated shape; semantic control changes the trajectory within that range rather than prescribing an absolute amplitude.
The first $96$ steps form $\tilde{x}$ and the last $96$ form $\tilde{y}$.

\section{Supplementary Experiments}
\label{sec:supplementary_experiments}

\subsection{Visual-Intermediate Fidelity}
\label{sec:supp_visual_fidelity}

We detail the generation-fidelity evaluation in Section~\ref{sec:exp_multimodal_generation}, report complementary diagnostics, and test the visual pathway with a fixed generator.

\subsubsection{Evaluation Protocol and Metric Definitions}
\label{sec:visual_fidelity_metrics}

The main comparison uses 4,450 paired, single-channel, 192-step windows across eight datasets.
The pathways receive identical \texttt{gpt-5.6-luna} semantic records and are paired by time-series ID.
Direct asks \texttt{gpt-5.6-luna} for a numerical array; VisualBridge renders the record with \texttt{gpt-image-2} and extracts the curve.
Both use records alone, without source-image conditioning, semantic edits, or VAE sampling.
The paired evaluation set contains complete windows from ETTh1 (126), ETTh2 (126), ETTm1 (504), ETTm2 (504), Exchange (56), and Weather (1,134).
Electricity and Traffic each use a fixed variable-stratified sample of 1,000 complete windows to cover their channels evenly.
Incomplete final windows are excluded.
The cycle metrics use 126, 124, 426, 458, 31, 547, 996, and 1,000 source-eligible windows in the same dataset order.

\paragraph{Metric definitions.}
For paired window $i$, let $\mathbf{x}_i^{\mathrm{src}}$ and
$\mathbf{x}_i^{\mathrm{gen}}$ denote the length-$L$ source and generated series.
We use the following average over an eligible set $\mathcal{I}$:
\[
    \langle u_i\rangle_{\mathcal{I}}
    =\frac{1}{|\mathcal{I}|}\sum_{i\in\mathcal{I}}u_i.
\]
An omitted subscript denotes all paired windows. To compare temporal shape across value scales, we robust-normalize each series:
\[
    \mathbf{z}(\mathbf{x})
    =\frac{\mathbf{x}-\operatorname{median}(\mathbf{x})}{R(\mathbf{x})},
    \qquad
    R(\mathbf{x})=Q_{0.95}(\mathbf{x})-Q_{0.05}(\mathbf{x}).
\]
For a zero quantile range, we use the peak-to-peak range, or one for a constant series.
Both pathways use the same deterministic decoders.
Trend and periodicity labels test agreement with the semantic record; the remaining metrics assess preservation of the source's temporal structure beyond those coarse attributes.
Together, they evaluate whether generated variations retain useful temporal properties without requiring pointwise reproduction of the source.
\begin{itemize}
    \item \textbf{Trend accuracy (Tr-Acc.)} compares the semantic target $y_i^{\mathrm{src}}$ with the generated trend label $y_i^{\mathrm{gen}}$ over all windows:
    \[
        \mathrm{Tr\text{-}Acc.}
        =\left\langle\mathbf{1}\!\left[y_i^{\mathrm{gen}}=y_i^{\mathrm{src}}\right]\right\rangle.
    \]
    The label set $\mathcal{C}_{\mathrm{tr}}$ contains decreasing, stable, increasing, and non-monotonic trends. The decoder uses linear displacement, monotonicity, and residual variation after smoothing.

    \item \textbf{Trend macro-F1 (Tr-F1)} gives each trend class equal weight on the same predictions:
    \[
        \mathrm{Tr\text{-}F1}
        =\frac{1}{|\mathcal{C}_{\mathrm{tr}}|}
        \sum_{c\in\mathcal{C}_{\mathrm{tr}}}
        \frac{2\mathrm{TP}_c}{2\mathrm{TP}_c+\mathrm{FP}_c+\mathrm{FN}_c}.
    \]
    Here, $\mathrm{TP}_c$, $\mathrm{FP}_c$, and $\mathrm{FN}_c$ count true positives, false positives, and false negatives for class $c$; an undefined class score is set to zero.

    \item \textbf{Periodicity F1 (Per-F1)} compares the semantic seasonality flag with the presence of at least two detected peaks in the generated series:
    \[
        \mathrm{Per\text{-}F1}
        =\frac{2\mathrm{TP}}{2\mathrm{TP}+\mathrm{FP}+\mathrm{FN}}.
    \]
    Peak counts $K_i^{\mathrm{src}}$ and $K_i^{\mathrm{gen}}$ are obtained after smoothing, using a common prominence and minimum-separation rule.

    \item \textbf{Cycle accuracy@$\pm1$ (Cyc-Acc.)} measures whether the generated peak count is within one of the source count:
    \[
        \mathrm{Cyc\text{-}Acc.}
        =\left\langle\mathbf{1}\!\left[
        |K_i^{\mathrm{gen}}-K_i^{\mathrm{src}}|\leq1
        \right]\right\rangle_{\mathcal{I}_{\mathrm{cyc}}},
        \qquad
        \mathcal{I}_{\mathrm{cyc}}=\{i:K_i^{\mathrm{src}}\geq2\}.
    \]
    Eligibility depends only on the source series, so both pathways are evaluated on the same cyclic windows.

    \item \textbf{Cycle-count MAE (Cyc-MAE)} measures the magnitude of the count error on the same cyclic windows:
    \[
        \mathrm{Cyc\text{-}MAE}
        =\left\langle|K_i^{\mathrm{gen}}-K_i^{\mathrm{src}}|
        \right\rangle_{\mathcal{I}_{\mathrm{cyc}}}.
    \]

    \item \textbf{Peak normalized location error (Peak-NLE)} measures displacement of the largest peak after Gaussian smoothing:
    \[
        p_i=\arg\max_t(G_{\sigma=2}*\mathbf{x}_i)_t,
    \]
    \[
        \mathrm{Peak\text{-}NLE}
        =\left\langle\frac{|p_i^{\mathrm{gen}}-p_i^{\mathrm{src}}|}{L-1}
        \right\rangle_{\mathcal{I}_{\mathrm{peak}}}.
    \]
    The set $\mathcal{I}_{\mathrm{peak}}$ excludes constant source windows.

    \item \textbf{Dominant-frequency error (DF-Err.)} compares the strongest frequencies in the unit-sum Hann periodogram $P_i(f)$ of each detrended, normalized series:
    \[
        f_i=\arg\max_{f\in[1/96,1/6]}P_i(f),
    \]
    \[
        \mathrm{DF\text{-}Err.}
        =\left\langle|f_i^{\mathrm{gen}}-f_i^{\mathrm{src}}|\right\rangle.
    \]

    \item \textbf{Turning-point-rate error (TPR-Err.)} measures differences in the frequency of local direction changes. Let $\mathbf{s}_i$ be the normalized series after second-order, nine-point Savitzky--Golay smoothing:
    \[
        r_{\mathrm{tp},i}=\frac{1}{L-2}\sum_{t=2}^{L-1}
        \mathbf{1}\!\left[
        \operatorname{sgn}(\Delta s_{i,t})
        \operatorname{sgn}(\Delta s_{i,t-1})<0\right],
    \]
    \[
        \mathrm{TPR\text{-}Err.}
        =\left\langle|r_{\mathrm{tp},i}^{\mathrm{gen}}-
        r_{\mathrm{tp},i}^{\mathrm{src}}|\right\rangle.
    \]
    A zero difference inherits the preceding nonzero sign.

    \item \textbf{Permutation-entropy error (PE-Err.)} compares local ordinal-pattern complexity. For the six order-three patterns $a\in S_3$ with probabilities $\pi_{i,a}$,
    \[
        H_{\mathrm{perm},i}
        =-\frac{1}{\log(3!)}\sum_{a\in S_3}\pi_{i,a}\log\pi_{i,a},
    \]
    \[
        \mathrm{PE\text{-}Err.}
        =\left\langle|H_{\mathrm{perm},i}^{\mathrm{gen}}-
        H_{\mathrm{perm},i}^{\mathrm{src}}|\right\rangle.
    \]

    \item \textbf{Increment Wasserstein distance (Inc.-$W_1$)} compares the empirical distributions of normalized first differences:
    \[
        \mathrm{Inc.\text{-}}W_1
        =\left\langle W_1(\Delta\mathbf{z}_i^{\mathrm{gen}},
        \Delta\mathbf{z}_i^{\mathrm{src}})\right\rangle.
    \]

    \item \textbf{Normalized four-band power $\ell_1$ distance (BP-$\ell_1$)} compares spectral power across four fixed frequency bands:
    \[
        (B_1,B_2,B_3,B_4)=([0,.05),[.05,.15),[.15,.30),[.30,.51)),
    \]
    \[
        q_{i,b}=\sum_{f\in B_b}P_i(f),
        \qquad
        \mathrm{BP\text{-}}\ell_1
        =\left\langle\frac{1}{2}\left\|\mathbf{q}_i^{\mathrm{gen}}-
        \mathbf{q}_i^{\mathrm{src}}\right\|_1\right\rangle.
    \]
\end{itemize}

\paragraph{Decoder settings.}
Trend decoding uses second-order, 19-point Savitzky--Golay smoothing.
A stable trend has absolute standardized displacement at most $0.35$ and first-difference roughness at most $0.003$.
Increasing and decreasing trends require displacement magnitude at least $0.70$, absolute Spearman correlation with time at least $0.75$, and residual standard-deviation ratio at most $0.65$; other cases are non-monotonic.
Peak counting uses second-order, 21-point smoothing, prominence $0.15$, and minimum separation $10$.
These rules are shared across datasets and generation pathways.

\subsubsection{Complementary Diagnostics}
\label{sec:visual_fidelity_diagnostics}

Table~\ref{tab:visual_fidelity_diagnostics} complements Table~\ref{tab:direct_vb_per_dataset} with trend agreement, cycle-count tolerance, increment distributions, and spectral power allocation.
The evaluation windows and generation conditions are unchanged.

\begin{table*}[t]
\centering
\caption{\textbf{Complementary semantic and structural diagnostics.}
Direct (D) and VisualBridge (VB) use the same paired windows as Table~\ref{tab:direct_vb_per_dataset}.}
\label{tab:visual_fidelity_diagnostics}
\setlength{\tabcolsep}{2pt}
\begin{small}
\resizebox{\textwidth}{!}{%
\begin{tabular}{lcccccccccccccccc}
\toprule
Metric
& \multicolumn{2}{c}{ETTh1}
& \multicolumn{2}{c}{ETTh2}
& \multicolumn{2}{c}{ETTm1}
& \multicolumn{2}{c}{ETTm2}
& \multicolumn{2}{c}{Exchange}
& \multicolumn{2}{c}{Weather}
& \multicolumn{2}{c}{Electricity}
& \multicolumn{2}{c}{Traffic} \\
\cmidrule(lr){2-3}\cmidrule(lr){4-5}\cmidrule(lr){6-7}\cmidrule(lr){8-9}\cmidrule(lr){10-11}\cmidrule(lr){12-13}\cmidrule(lr){14-15}\cmidrule(lr){16-17}
& D & VB & D & VB & D & VB & D & VB & D & VB & D & VB & D & VB & D & VB \\
\midrule
Tr-Acc. $\uparrow$
& 84.9\% & \textbf{88.9\%} & \textbf{77.0\%} & 75.4\%
& \textbf{93.5\%} & 91.7\% & \textbf{87.5\%} & 86.5\%
& \textbf{76.8\%} & 71.4\% & \textbf{84.7\%} & \textbf{84.7\%}
& \textbf{60.6\%} & 60.5\% & \textbf{79.5\%} & 77.6\% \\
Tr-F1 $\uparrow$
& 46.9\% & \textbf{58.4\%} & \textbf{51.8\%} & 48.7\%
& 45.6\% & \textbf{47.2\%} & \textbf{48.1\%} & 41.2\%
& \textbf{56.3\%} & 53.9\% & \textbf{69.7\%} & 67.0\%
& 35.2\% & \textbf{46.1\%} & \textbf{32.1\%} & 28.1\% \\
Cyc-Acc. $\uparrow$
& 11.9\% & \textbf{27.0\%} & 16.1\% & \textbf{29.8\%}
& 50.2\% & \textbf{52.1\%} & 53.1\% & \textbf{57.0\%}
& \textbf{61.3\%} & \textbf{61.3\%} & 67.8\% & \textbf{69.7\%}
& 35.5\% & \textbf{36.1\%} & \textbf{28.4\%} & 20.9\% \\
Inc.-$W_1$ $\downarrow$
& 0.1104 & \textbf{0.0583} & 0.1483 & \textbf{0.0906}
& 0.0454 & \textbf{0.0439} & \textbf{0.1101} & 0.1118
& 0.0348 & \textbf{0.0302} & \textbf{0.0750} & 0.1197
& 0.1318 & \textbf{0.0859} & 0.1163 & \textbf{0.0714} \\
BP-$\ell_1$ $\downarrow$
& 0.3464 & \textbf{0.2535} & 0.3696 & \textbf{0.2831}
& 0.1049 & \textbf{0.1038} & \textbf{0.1626} & 0.1789
& 0.1116 & \textbf{0.0783} & 0.0899 & \textbf{0.0892}
& 0.5296 & \textbf{0.4745} & 0.4424 & \textbf{0.3445} \\
\bottomrule
\end{tabular}%
}
\end{small}
\vspace{-8pt}
\end{table*}

VisualBridge generally improves cycle-count tolerance and spectral power allocation, complementing the main table's evidence of recurring-structure preservation.
Trend-label comparisons favor Direct on most datasets, while first-difference distribution errors favor different pathways on different datasets.
Together with Table~\ref{tab:direct_vb_per_dataset}, these results locate the visual pathway's main benefits in recurring structure and local shape.

\subsubsection{Same-Model Control for the Visual Pathway}
\label{sec:visual_same_model}
To isolate the contribution of the visual pathway, we fix the generator and vary only how a semantic record is converted to a sequence.
On 182 paired windows from Exchange and ETTh1, both branches use
\texttt{gemini-3.1-flash-image-preview} and receive identical semantic records.
The Direct branch requests a numerical array. The VisualBridge branch requests
a plot and applies deterministic curve extraction to obtain the array. This
design attributes the comparison to the visual intermediate rather than to a
difference between generation models.

We report both trend accuracy and macro-F1 because the trend classes are imbalanced: the 182 targets contain 150 non-monotonic, 17 increasing, 12 decreasing, and 3 stable windows.
Macro-F1 gives these classes equal weight, complementing accuracy over all windows.

\begin{table}[!htbp]
\centering
\caption{\textbf{Same-model comparison between direct
numerical generation and VisualBridge.} ETTh1 and Exchange contain 126 and 56
paired windows, respectively. Both
pathways use \texttt{gemini-3.1-flash-image-preview} with identical semantic
records, so the comparison isolates the output pathway. The cycle metrics use
126 ETTh1 and 31 Exchange windows.}
\label{tab:gemini_same_model_pathway}
\setlength{\tabcolsep}{6pt}
\begin{small}
\begin{tabular}{lcccc}
\toprule
Metric
& \multicolumn{2}{c}{ETTh1}
& \multicolumn{2}{c}{Exchange} \\
\cmidrule(lr){2-3}\cmidrule(lr){4-5}
& Direct & VisualBridge & Direct & VisualBridge \\
\midrule
Tr-Acc. $\uparrow$
& 74.6\% & \textbf{87.3\%}
& 83.9\% & \textbf{85.7\%} \\
Tr-F1 $\uparrow$
& 32.6\% & \textbf{46.3\%}
& \textbf{84.5\%} & 64.3\% \\
Per-F1 $\uparrow$
& 71.6\% & \textbf{84.3\%}
& 12.5\% & \textbf{21.1\%} \\
Cyc-Acc. $\uparrow$
& 15.9\% & \textbf{33.3\%}
& 41.9\% & \textbf{51.6\%} \\
Cyc-MAE $\downarrow$
& 3.4683 & \textbf{2.6667}
& 1.6774 & \textbf{1.5806} \\
Peak-NLE $\downarrow$
& 0.2914 & \textbf{0.2531}
& \textbf{0.0925} & 0.1155 \\
\bottomrule
\end{tabular}
\end{small}
\vspace{-8pt}
\end{table}

VisualBridge improves all six metrics on ETTh1 and improves trend accuracy, periodicity F1, and both cycle metrics on Exchange.
Direct retains higher trend macro-F1 and lower peak-location error on Exchange.
With the generator and semantic input fixed, the recurring-structure gains on both datasets support the visual intermediate within the same model.

\subsection{Semantic-Action Realization}
\label{sec:supp_action_realization}

We measure whether semantic edits produce the requested numerical changes in paired generated sequences.
An edit succeeds when the corresponding statistic changes in the requested direction relative to matched identity: absolute linear slope for trend, the dominant spectral power ratio for seasonality, and first-difference standard deviation for volatility.
Appendix~\ref{sec:multimodal_control_impl} defines these statistics.
Table~\ref{tab:action_realization} reports success rates over paired generations, with each channel-level request counted once.
All evaluated records contain finite extracted curves.
All three attributes change in the requested direction in most evaluated pairs across all eight datasets, with volatility having the highest realization rate on six datasets.
The visual pathway thus provides a practical means of realizing the policy's semantic actions.
RL then learns which of these realizable changes benefit forecasting, using the matched task reward rather than directional agreement alone.

\begin{table}[!htbp]
\centering
\caption{\textbf{Realization of semantic actions in matched generations.} Cells report the percentage of paired generations in which the requested attribute changes in the intended direction.}
\label{tab:action_realization}
\begin{small}
\setlength{\tabcolsep}{6pt}
\begin{tabular}{lccc}
\toprule
Dataset & Trend & Seasonality & Volatility \\
\midrule
ETTh1 & 85.9 & 79.2 & 100.0 \\
ETTh2 & 89.4 & 81.5 & 77.3 \\
ETTm1 & 94.2 & 74.5 & 99.1 \\
ETTm2 & 78.4 & 85.2 & 92.4 \\
Weather & 82.8 & 73.6 & 100.0 \\
Electricity & 83.4 & 80.0 & 76.9 \\
Traffic & 93.1 & 98.1 & 99.3 \\
Exchange & 85.4 & 88.5 & 100.0 \\
\bottomrule
\end{tabular}
\end{small}
\end{table}

\subsection{Matched Identity Control}
\label{sec:supp_reward_reference}

We compare policies whose task utility uses either matched identity error or original-only proxy error as the numerator of Eq.~\ref{eq:semantic_downstream_utility}.
In the original-only condition, this substitution is made before computing the reward and its utility-dependent auxiliary cap; identity retains reward zero in the action group.
Both conditions retain the same reliability checks, downside penalties, admission criteria, and action and image-call budgets.
Each policy is trained independently and used to construct augmentation for the same final forecasting protocol.
Table~\ref{tab:reward_reference} shows lower forecasting MSE with the matched identity reference on every dataset.
The comparison evaluates the reward reference within the complete learning and collection procedure: it affects reward calibration, group advantages relative to the fixed identity control, and positive-reward admission.
The results support using no-edit visual generation as the reference for selecting forecasting-relevant semantic variations.

\begin{table}[!htbp]
\centering
\caption{\textbf{Effect of the task-utility reference.} Test MSE of final forecasters trained with augmentation from each policy.}
\label{tab:reward_reference}
\begin{small}
\setlength{\tabcolsep}{9pt}
\begin{tabular}{lcc}
\toprule
Dataset & Matched identity & Original-only reference \\
\midrule
ETTh1 & \textbf{0.459}{\scriptsize$\pm$0.001} & 0.466{\scriptsize$\pm$0.001} \\
ETTh2 & \textbf{0.292}{\scriptsize$\pm$0.001} & 0.299{\scriptsize$\pm$0.001} \\
ETTm1 & \textbf{0.415}{\scriptsize$\pm$0.002} & 0.421{\scriptsize$\pm$0.002} \\
ETTm2 & \textbf{0.186}{\scriptsize$\pm$0.001} & 0.188{\scriptsize$\pm$0.001} \\
Weather & \textbf{0.173}{\scriptsize$\pm$0.001} & 0.174{\scriptsize$\pm$0.001} \\
Electricity & \textbf{0.171}{\scriptsize$\pm$0.000} & 0.172{\scriptsize$\pm$0.000} \\
Traffic & \textbf{0.425}{\scriptsize$\pm$0.002} & 0.439{\scriptsize$\pm$0.002} \\
Exchange & \textbf{0.085}{\scriptsize$\pm$0.000} & 0.088{\scriptsize$\pm$0.000} \\
\bottomrule
\end{tabular}
\end{small}
\end{table}

\subsection{Transfer to Other Forecasting Backbones}
\label{sec:supp_backbone_transfer}

We evaluate whether augmentation constructed with iTransformer also benefits other forecasting architectures.
PatchTST~\citep{nie2022time} and DLinear~\citep{zeng2023transformers} replace only the final forecaster on ETTh1, ETTm1, Weather, and Exchange.
The semantic action sampler, VAE, and augmented samples remain fixed, so neither augmentation generation nor selection is adapted to the receiving backbone.

For each backbone, Original uses real few-shot windows, while VisualBridge adds the same augmented set used for iTransformer.
Both conditions share data partitions, normalization, input and prediction lengths, and backbone-specific training settings.
Checkpoints are selected by validation MSE and evaluated on the test split.
Table~\ref{tab:backbone_transfer} shows lower MAE and MSE for both backbones on every evaluated dataset, demonstrating that the fixed augmented set transfers beyond the forecaster used to construct it.

\begin{table}[!htbp]
\centering
\caption{\textbf{Augmentation transfer across forecasting backbones.} Original uses real data only; VisualBridge adds a fixed augmented set.}
\label{tab:backbone_transfer}
\begin{small}
\setlength{\tabcolsep}{6pt}
\begin{tabular}{llcccc}
\toprule
\multirow{2}{*}{Backbone} & \multirow{2}{*}{Dataset} & \multicolumn{2}{c}{Original} & \multicolumn{2}{c}{VisualBridge} \\
\cmidrule(lr){3-4}\cmidrule(lr){5-6}
& & MAE & MSE & MAE & MSE \\
\midrule
PatchTST & ETTh1 & 0.484 & 0.512 & \textbf{0.465} & \textbf{0.496} \\
PatchTST & ETTm1 & 0.405 & 0.416 & \textbf{0.389} & \textbf{0.397} \\
PatchTST & Weather & 0.213 & 0.178 & \textbf{0.206} & \textbf{0.174} \\
PatchTST & Exchange & 0.211 & 0.091 & \textbf{0.207} & \textbf{0.088} \\
\midrule
DLinear & ETTh1 & 0.460 & 0.469 & \textbf{0.445} & \textbf{0.451} \\
DLinear & ETTm1 & 0.418 & 0.429 & \textbf{0.411} & \textbf{0.423} \\
DLinear & Weather & 0.254 & 0.200 & \textbf{0.243} & \textbf{0.197} \\
DLinear & Exchange & 0.211 & 0.092 & \textbf{0.209} & \textbf{0.086} \\
\bottomrule
\end{tabular}
\end{small}
\end{table}

\section{Dataset Details and Forecasting Protocol}
\label{sec:dataset_details}

The experimental protocol is described in Section~\ref{sec:exp_setup}.
Here we report the dataset statistics and the exact partitions used in the few-shot setting.

\begin{table}[ht]
    \centering
    \caption{\textbf{Dataset statistics and partitions.}
    Counts refer to windows with $96$ input and $96$ prediction steps; partitions are train/validation/test.
    Few-shot training uses the earliest $20\%$ of training windows.
    }
    \label{tab:dataset_statistics}
    \setlength{\tabcolsep}{4pt}
    \begin{small}
    \resizebox{\linewidth}{!}{
    \begin{tabular}{lcccc}
        \toprule
        Dataset & Features & Total Windows & Standard Partition & Few-shot Partition \\
        \midrule
        ETTh1 / ETTh2 & 7 & 14019 & 8449 / 2785 / 2785 & 1689 / 2785 / 2785 \\
        ETTm1 / ETTm2 & 7 & 57219 & 34369 / 11425 / 11425 & 6873 / 11425 / 11425 \\
        Traffic & 862 & 17163 & 12089 / 1661 / 3413 & 2417 / 1661 / 3413 \\
        Electricity & 321 & 25923 & 18221 / 2537 / 5165 & 3644 / 2537 / 5165 \\
        Weather & 21 & 52315 & 36696 / 5175 / 10444 & 7339 / 5175 / 10444 \\
        Exchange & 8 & 7207 & 5120 / 665 / 1422 & 1024 / 665 / 1422 \\
        \bottomrule
    \end{tabular}
    }
    \end{small}
\end{table}

Windows have stride one, and the few-shot count is rounded down.
Validation and test windows include the preceding $96$ time steps as context; all targets remain within the corresponding interval.
Per-channel means and standard deviations are computed once from the standard training interval and reused by every few-shot baseline, VisualBridge, and the full-data reference.
This common transformation keeps the input scale and reported errors comparable.
In the few-shot experiments, the remaining training windows supply only these normalization statistics, not fitting examples or augmentation sources.
Semantic generation, policy rewards, candidate selection, and VAE fitting use the few-shot training data.
Validation and test observations do not enter normalization or augmentation construction: validation selects final forecasting checkpoints, and the test split is used only for final evaluation.
The forecasting protocol is specified in Appendix~\ref{sec:final_forecast_impl}.

\section{Implementation Details}
\label{sec:implementation_details}

This section specifies semantic action sampler training, augmentation-anchor collection, VAE sampling, and final forecasting.
Semantic QA and image-to-array conversion follow Appendices~\ref{sec:qa_protocol} and~\ref{sec:image_to_array}.
The data-use protocol follows Appendix~\ref{sec:dataset_details}.
Tables~\ref{tab:implementation_hyperparams} and~\ref{tab:policy_hyperparams} collect the model and RL settings, respectively.

\begin{table}[!t]
    \centering
    \caption{\textbf{Model and forecasting settings.} All final forecasters use seed $2023$; repeated results vary the augmented data.}
    \label{tab:implementation_hyperparams}
    \setlength{\tabcolsep}{4pt}
    \begin{small}
    \renewcommand{\arraystretch}{0.90}
    \resizebox{\linewidth}{!}{%
    \begin{tabular}{lll}
        \toprule
        Component & Setting & Value \\
        \midrule
        \multicolumn{3}{l}{\textit{Forecasting and visual generation}} \\
        Forecaster & Backbone and objective & iTransformer; $L_1$ loss \\
        Forecaster & Hidden dimensions & $d_{\mathrm{model}}=128$, $d_{\mathrm{ff}}=128$ \\
        Forecaster & Encoder architecture & 2 layers; 8 attention heads \\
        Forecaster & Optimization & Adam; learning rate $10^{-4}$ \\
        Semantic QA & Model & gpt-5.6-luna \\
        Plot generation & Model & gpt-image-2 \\
        \midrule
        \multicolumn{3}{l}{\textit{Temporal VAE and posterior sampling}} \\
        Architecture & Dimensions & Latent $64$; hidden $256$ \\
        Architecture & Convolutions and dropout & 3 layers; dropout $0.1$ \\
        Optimization & Optimizer and batch size & AdamW; batch size $128$ \\
        Optimization & Learning rates & Pretraining $10^{-3}$; refinement $10^{-4}$ \\
        VAE objective & Difference and KL weights & Difference $0.3$; maximum KL $10^{-3}$ \\
        Decoder & Task weight and sampling scale & Task weight $0.1$; $\gamma=0.5$ \\
        Candidate selection & Samples per anchor & 5 draws; retain at most 1 \\
        Candidate checks & Scale-ratio bounds & $[0.75,1.25]$ \\
        Candidate checks & Forecaster-loss-ratio bounds & $[0.5,1.25]$ \\
        \midrule
        \multicolumn{3}{l}{\textit{Original training budget}} \\
        Dataset & Epochs & Batch size \\
        ETTh1, Exchange & $10$ & $32$ \\
        ETTm1, Weather & $20$ & $32$ \\
        ETTh2, ETTm2 & $30$ & $32$ \\
        Electricity & $100$ & $16$ \\
        Traffic & $100$ & $2$ \\
        \bottomrule
    \end{tabular}
    }
    \end{small}
\end{table}

\begin{table}[!t]
    \centering
    \caption{\textbf{Semantic action sampler and reward settings.}}
    \label{tab:policy_hyperparams}
    \setlength{\tabcolsep}{4pt}
    \begin{small}
    \renewcommand{\arraystretch}{0.90}
    \resizebox{\linewidth}{!}{%
    \begin{tabular}{lll}
        \toprule
        Component & Setting & Value \\
        \midrule
        \multicolumn{3}{l}{\textit{Action sampling and policy updates}} \\
        Action sampling & Group size and exploration & 4 actions; mixture weight $0.20$ \\
        Candidate blending & Generated-sequence weight $\rho$ & $0.10$ \\
        Policy update & Optimizer and steps per group & Adam; 12 steps \\
        Policy update & Learning rate & $0.03$ \\
        Policy update & Clipping and regularization & Clip $0.20$; KL $0.08$; entropy $0.02$ \\
        Frozen-policy collection & Log-probability weight $\kappa$ & $0.02$ \\
        \midrule
        \multicolumn{3}{l}{\textit{Proxy forecasting and task reward}} \\
        Reward folds & Count and purge gap & 3 folds; 191 window starts \\
        Reward folds & End fractions; probe fraction & $0.68,0.82,0.96$; $0.08$ \\
        Proxy forecaster & Update budget and gradient clip & 5 epochs; clip norm $1.0$ \\
        Proxy comparison & Generated supervision & Loss weight $0.5$; raw weights $[0.25,1.0]$ \\
        Task reward & Metric weights & MSE $1.0$; MAE $0.5$ \\
        Probe summary & Temporal weights & Best $0.7$; tail $0.3$ (last 3 epochs) \\
        Reward aggregation & Scale and fold penalty & Task scale $10$; fold variation $0.5$ \\
        Reward penalties & Downside; instability; missing fraction & $4$; $0.25$; $0.20$ \\
        \midrule
        \multicolumn{3}{l}{\textit{Reliability and acceptance}} \\
        Reliability & Auxiliary weights & Semantic $0.05$; visual $0.02$; forecaster $0.02$ \\
        Reliability & Auxiliary relative cap & $0.25$ \\
        Acceptance & Minimum valid fraction; invalid reward & $0.5$; $-10$ \\
        Downside bound & Identity tolerance; excess limit & $0.002$; $0.001$ \\
        Direction score & Margins (trend, seasonality, volatility) & $0.03,0.02,0.05$ \\
        Direction score & Scales (trend, seasonality, volatility) & $0.15,0.08,0.15$ \\
        Curve quality & Minimum score; maximum identity gap & $0.65$; $0.15$ \\
        \bottomrule
    \end{tabular}
    }
    \end{small}
\end{table}

\subsection{Semantic Action Sampler and Action Construction}
\label{sec:semantic_policy_impl}

The context $h_i$ has eight numerical and thirteen semantic entries, computed over the controlled channels (all channels when no subset is selected).
The numerical entries are a bias, mean absolute linear slope, spectral seasonality, first-difference volatility, window and difference standard deviations, and the frozen original-only forecaster's MAE and $\log(1+\mathrm{MAE})$.

The semantic entries record channel proportions for the three trend-strength levels, seasonality existence and its three strength levels, and the three volatility levels.
The remaining entries are the channel-averaged counts of change points, anomalies, and segments, divided by $4$, $4$, and $6$, respectively, and capped at $1$.
The semantic action sampler uses a linear softmax policy, restricted to admissible actions:
\[
    \pi_{\theta}(u\mid h)=\operatorname{softmax}(Wh+b)_u.
\]
Training uses the batch-mean context because each action receives a shared forecasting reward for that batch.
With a linear policy, its logits at the batch mean equal the mean of the per-window logits:
\[
    Wh_{\mathcal B}+b=\frac{1}{|\mathcal B|}\sum_{i\in\mathcal B}(Wh_i+b).
\]
Collection applies this same learned mapping to a single window; neither the descriptor definition nor the policy parameters change.

The seven actions comprise identity and one-step strengthening or weakening of trend strength, seasonality strength, and volatility level.
Trend strength follows \texttt{weak}, \texttt{medium}, \texttt{strong}; volatility follows \texttt{low}, \texttt{medium}, \texttt{high}.
Seasonality adds \texttt{none} below its three strength levels, so weakening \texttt{weak} seasonality removes it and strengthening \texttt{none} introduces \texttt{weak} seasonality.
Steps are clipped at the ends of each ordering, and trend direction remains unchanged.
An unknown trend or volatility level maps to \texttt{medium}; absent or unknown seasonality starts from \texttt{none}.
Volatility edits also step the segment-level volatility labels.
Schema repair clears the type and period when seasonality is absent; newly introduced seasonality uses \texttt{irregular} and a \texttt{medium} period hint.
The summary is regenerated deterministically from the edited fields without another model call; event indices and segment boundaries remain fixed.
For each window, an edit is admissible when it changes at least one controlled channel; ineffective channel records remain unchanged.
Training uses actions admissible for every window in its source batch.

Each group includes identity; the remaining actions are sampled without replacement from a mixture of the learned policy and uniform exploration over admissible edits.
If fewer than three non-identity edits are admissible, the group contains all available edits; a batch without an admissible edit is excluded.
The policy carries its learned preferences across rounds, so forecasting feedback changes subsequent proposals.

\subsection{Matched Generation and Candidate Checks}
\label{sec:multimodal_control_impl}

Every action and its identity control use the same source reference plots, image model, output settings, extractor, and scale calibration.
Each request generates one independent image per controlled channel; requests from different training rounds and the collection stage are kept separate.

For raw recovered candidates, the directional change relative to identity is
\[
    e=d_u\log\frac{m_u+\epsilon}{m_0+\epsilon}.
\]
Here, $m_u$ and $m_0$ are the edited and identity attribute statistics, and $d_u\in\{-1,1\}$ denotes weakening or strengthening.
Trend strength is the mean absolute linear slope across controlled channels, using time coordinates in $[-1,1]$.
Seasonality strength is the channel-averaged ratio of peak to total nonzero-frequency power after linear detrending; volatility is the standard deviation of first differences.
These statistics test edit direction, while local segments and events remain part of the semantic generation conditions.
The requested JSON edit must be effective, and the generated image must change without reversing the requested numerical direction.

The forecasting candidate blends the recovered sequence with its source:
\[
    c_i^u=(1-\rho)s_i+\rho\tilde{s}_i^u.
\]
The same blending coefficient is applied to identity.
With $\rho=0.10$, an augmentation anchor contains $90\%$ of its source sequence and $10\%$ of the recovered generation, retaining source structure while incorporating the selected semantic variation.
Both its input and target are taken directly from this blended sequence.
We check finite values, training-derived value and variation bounds, curve-extraction quality, and the ratio of its frozen-forecaster MAE to that of the source window.
A source window enters the paired downstream comparison only when its candidates pass the checks in both branches.

Both proxy branches use the same generated-loss coefficient and reliability-weighting rule, evaluated separately for their candidates:
\begin{equation}
    \omega=\operatorname{clip}\!\left(e^{-|\log r|}\bar q\max(1/4,c),\omega_{\min},\omega_{\max}\right).
\end{equation}
Here, $r$ is the source-relative frozen-forecaster MAE ratio and $\bar q$ is the mean raw curve-extraction score over controlled channels. The directional weight is
\[
    c=\begin{cases}
        (1+S)/2, & \text{for a semantic edit},\\
        1, & \text{for identity}.
    \end{cases}
\]
Appendix~\ref{sec:reward_grpo_impl} defines $S$ and the curve-extraction score.
Before proxy training, the clipped weights are divided by their mean within each branch's retained candidate set, reducing the relative influence of less reliable candidates.
Edited and identity candidates use the same retained source indices within each pair but receive their own weights; different actions can retain different subsets.
The reward therefore evaluates each action together with the prescribed filtering and weighting, as used to construct augmentation.
The same original-only forecaster provides context features, candidate checks, VAE consistency loss, and VAE candidate selection; it remains frozen throughout these stages.
The proxy models used for reward computation and the final forecaster are trained separately from scratch.

For a channel budget $C_{\mathrm{edit}}$, editable channels are ranked by the frozen-forecaster error of their aligned no-edit generation relative to the original, with the fraction of admissible edits used as a secondary preference.
We keep the first $\min(C,C_{\mathrm{edit}})$ channels and use this same set throughout search and collection.
Uncontrolled channels retain aligned no-edit generated values in both branches.

\subsection{Downstream Reward and GRPO}
\label{sec:reward_grpo_impl}

Chronological reward folds are formed entirely inside the few-shot training split.
Each fold contains an update prefix and a later probe block, separated by $L_x+L_y-1$ window starts so that their complete forecasting windows do not overlap.
Generation source windows are restricted to the earliest update prefix during policy training.
For each source batch and fold, original-only, matched identity, and edited branches share the proxy-forecaster initialization, original-data batches, and update budget.
Each proxy is initialized from scratch with the same seed and fitted only on that fold's update prefix; probe windows never enter its gradient updates.
The frozen original-only forecaster supplies context features and candidate checks from the few-shot training data, but its weights are not used to initialize the proxies.
Probe blocks supply internal policy-training feedback; the official validation and test splits remain outside this procedure.

For metric $m\in\{\mathrm{MSE},\mathrm{MAE}\}$, the probe error $\ell_m^u$ combines its best value and the mean of its final values, accounting for both attained accuracy and late-update behavior.
We omit batch and fold indices below.
The task term scales the paired forecasting utility from Eq.~\ref{eq:semantic_downstream_utility}:
\[
    G=\eta q(u).
\]

The downside penalty measures deterioration beyond that caused by identity generation:
\[
    d_m^u=\left[\log\frac{\ell_m^u}{\ell_m^{\mathrm{ori}}}\right]_+,
    \qquad
    D=\sum_m w_m\left[d_m^u-d_m^0-\tau\right]_+.
\]
Here, $[\cdot]_+$ is the positive part, $\tau$ is a tolerance, and $w_m$ uses the same metric weights as the task utility.
The instability penalty $J$ is the positive excess over identity in the coefficient of variation of the final probe MSE values.
The fold reward is
\begin{equation}
    r=G+A-\lambda_D D-\lambda_J J-\lambda_v(1-v)|G|,
    \label{eq:fold_semantic_reward}
\end{equation}
where $v$ is the fraction of source windows with valid paired candidates.
A fold with insufficient valid paired candidates receives a fixed invalid reward.
The action reward is the mean fold reward minus a weighted standard deviation across folds.

\paragraph{Auxiliary scores.}
The auxiliary term combines directional agreement $S$, change in curve quality $\Delta Q$, and forecasting consistency $F$:
\[
    A=\operatorname{clip}\!\left(
    \left\langle\lambda_S S+\lambda_Q\Delta Q+\lambda_F F\right\rangle_{\mathcal{B}},
    -c_{\mathrm{aux}}|G|,\ c_{\mathrm{aux}}|G|
    \right),\qquad 0\leq c_{\mathrm{aux}}<1.
\]
The cap keeps these reliability scores subordinate to forecasting utility: they cannot reverse the task term's sign before penalties.
Directional agreement is a smooth score of the requested numerical change:
\[
    S=\tanh\!\left(\frac{e-\delta_u}{t_u}\right).
\]
The margin $\delta_u$ and scale $t_u$ account for the different attribute statistics.
Forecasting consistency $F$ is triangular in the log forecaster-loss ratio: it peaks at one at the interval's geometric midpoint, reaches zero at its bounds, and decreases outside to a minimum of $-1$.
The loss-ratio interval and reward coefficients are collected in Tables~\ref{tab:implementation_hyperparams} and~\ref{tab:policy_hyperparams}.

\paragraph{Curve-extraction score.}
Image quality is evaluated at a common resolution using nearest-neighbor resizing.
The raw quality score $q_{\mathrm{img}}\in[0,1]$ combines horizontal coverage, occupied columns, gap length, boundary contact, and vertical ambiguity within a column.
It rewards broad, continuous traces and penalizes clipping or multiple plausible curve positions.
These checks concern curve extraction, not forecasting accuracy.

Acceptance requires a readable trace and limits quality loss relative to matched identity.
For the reward, two smooth transforms score the margins to these acceptance bounds; their edited-minus-identity difference defines $\Delta Q$.
This score assesses whether the requested edit remains reliably recoverable as a numerical sequence.

\paragraph{Policy updates.}
Each round draws a source batch, samples an action group, makes the matched image requests, evaluates the candidates on all reward folds, and updates the policy before the next round.
The evaluated rewards, advantages, and behavior probabilities remain fixed for 12 Adam steps on that group.
For each sampled non-identity action, its likelihood is conditioned on removing earlier draws from the exploration mixture.
The ratio $r_j$ compares this probability under the current policy with that under the policy that generated the group.
Using the advantage $\widehat A_j$ from Eq.~\ref{eq:semantic_grpo_advantage}, the clipped surrogate for sampled action $j$ is
\[
    \mathcal{L}_j^{\mathrm{clip}}
    =\min\!\left\{
    r_j\widehat A_j,\,
    \operatorname{clip}(r_j,1-\epsilon_c,1+\epsilon_c)\widehat A_j
    \right\}.
\]

We maximize the mean surrogate with a forward KL penalty and an entropy bonus:
\[
    \mathcal J=\frac{1}{K}\sum_{j=1}^{K}\mathcal L_j^{\mathrm{clip}}
    -\beta_{\mathrm{KL}}D_{\mathrm{KL}}(\pi_\theta\Vert U)
    +\beta_H H(\pi_\theta).
\]
Here, $K$ is the number of sampled non-identity edits and $U$ is uniform over their admissible set $\mathcal U$.
Both regularizers use the unmixed policy on this set. Since
\[
    D_{\mathrm{KL}}(\pi_\theta\Vert U)=\log|\mathcal U|-H(\pi_\theta),
\]
they jointly provide entropy regularization with weight $\beta_{\mathrm{KL}}+\beta_H$, maintaining exploration across rounds.
Identity has reward zero and enters advantage normalization, but not the sampled-action objective.
The updated policy determines the next round's action distribution.

\paragraph{Frozen-policy collection.}
After the fixed search budget is exhausted, let $\bar R(u)$ be the mean reward of an edit across its evaluated source batches.
An edit must have positive mean reward and mean paired utility.
It must also pass a matched evaluation with positive reward on every fold, non-negative MSE and combined gain over identity, and bounded incremental MSE downside relative to identity.
For each collection source window, let $\mathcal{U}_i^+$ contain eligible edits that remain admissible for its record.
The frozen policy selects
\begin{equation}
    u_i^*=\arg\max_{u\in\mathcal{U}_i^+}
    \left[\bar R(u)+\kappa\log\pi_{\theta^*}(u\mid h_i)\right].
    \label{eq:frozen_anchor_collection}
\end{equation}
Ties follow a fixed action order; windows with an empty eligible set are skipped.
The mean reward summarizes performance across evaluated batches, while the policy probability depends on the collection window's context.

New edited and matched identity requests are made, and the candidate checks are repeated without a policy update or a new downstream rollout.
The set $\mathcal{A}$ consists exclusively of accepted blended sequences from these fresh requests.
Collection windows remain within the few-shot training interval, although their source indices may have appeared during policy training.

\subsection{Temporal VAE Training and Sampling}
\label{sec:vae_expansion_impl}

The temporal convolutional VAE jointly encodes all channels of $s=[x,y]$ into a diagonal Gaussian posterior $q_{\phi}(z\mid s)$; $D_{\eta}$ decodes the complete multivariate window.
It models cross-channel and input--target relationships after channel-wise visual generation, then samples around the anchors selected through semantic control.

\paragraph{Joint reconstruction.}
Pretraining combines real few-shot windows, aligned no-edit generations, and augmentation anchors.
A later block within the few-shot training interval monitors reconstruction, separated from fitting windows by the same purge rule; generated samples follow their source windows' split.

The VAE minimizes $L_1$ reconstruction losses on values and first differences, plus a KL penalty toward a standard Gaussian prior.
The difference loss fits local increments rather than suppressing variation.
The KL weight is warmed up during pretraining.

\paragraph{Forecasting-consistent decoding.}
We then freeze the encoder and original-only forecaster $F_{\mathrm{ori}}$ and refine only the decoder on real training windows.
Value and first-difference reconstruction constrain deviation from these windows, while the forecasting-consistency loss is
\[
    \mathcal{L}_{\mathrm{cons}}
    =\left\lVert F_{\mathrm{ori}}(\hat{x})-\hat{y}\right\rVert_1.
\]
The decoder produces $[\hat{x},\hat{y}]$ jointly; this loss encourages input--target compatibility as measured by the frozen forecaster, while reconstruction retains the observed multivariate trajectory.

\paragraph{Anchor-based sampling.}
For each $b_i\in\mathcal{A}$, the frozen VAE decodes posterior draws by Eq.~\ref{eq:anchor_vae_sampling}.
Candidates must be finite and satisfy value bounds, anchor-relative overall and first-difference scale bounds, and a forecaster-loss-ratio band.
Only the valid candidate with the lowest forecaster MAE is retained, including its decoded input and target; anchors with no valid draw contribute no sample.
The resulting set contains at most $|\mathcal{A}|$ samples.
Directional checks evaluate the visual candidates before blending; the VAE stage then models and selects complete forecasting pairs around the resulting anchors.
The sample-matched comparison with w/o VAE in Section~\ref{sec:exp_ablation} evaluates this complete stage, including joint reconstruction, decoder refinement, posterior sampling, and candidate selection.

\subsection{Final Forecast Training}
\label{sec:final_forecast_impl}

The final iTransformer is trained from scratch with the policy and VAE frozen.
The forecasting seed is fixed at $2023$ across Original and all augmentation methods; the three repeated runs use independently generated augmentation sets.
Original and augmented losses are averaged within separate batches and combined by Eq.~\ref{eq:final_controlled_augmentation}.
If no augmented sample passes selection, training uses only real data.
For all final forecasting comparisons, the checkpoint is selected by the lowest validation MSE within the training budget.
The selected checkpoint is then evaluated on the test set for both MAE and MSE.

\paragraph{Computational cost.}
For $N$ source windows and $C$ channels, semantic extraction makes $NC$ requests, with each resulting record reused for subsequent edits.
With fixed window length, image resolution, and requests per window, visual generation and curve recovery scale linearly in the number of window--channel pairs.
Policy evaluation and VAE training add offline construction cost; controlled-channel selection bounds the number of edited image requests.
The augmented set is reusable across forecasting backbones, as evaluated in Appendix~\ref{sec:supp_backbone_transfer}.
The downstream architecture is unchanged and requires no semantic model, image generator, policy, or VAE at inference.

\end{document}